\documentclass[10pt,journal,compsoc]{IEEEtran}
\pdfoutput=1

\ifCLASSOPTIONcompsoc
  \usepackage[nocompress]{cite}
\else
  \usepackage{cite}
\fi

\ifCLASSINFOpdf
\else
\fi

\usepackage[utf8]{inputenc} 
\usepackage[T1]{fontenc}    
\usepackage{url}            
\usepackage{booktabs}       
\usepackage{amsfonts}       
\usepackage{nicefrac}       
\usepackage{microtype}      
\usepackage{amsmath, amssymb}
\usepackage{graphicx}
\usepackage{color}
\usepackage{bm}
\usepackage{caption}
\usepackage{multirow}
\usepackage{wrapfig}
\usepackage{tikz}
\usepackage{subcaption}
\usepackage{xcolor}

\usepackage[english]{babel}

\usepackage[colorlinks = true,
            linkcolor = red,
            urlcolor  = blue,
            citecolor = green,
            anchorcolor = red]{hyperref}

\newcommand{\ours}{Adaptive RPS-Net}

\begin{document}
\title{An Adaptive Random Path Selection Approach for Incremental Learning}

\author{Jathushan~Rajasegaran, 
        Munawar~Hayat,
        Salman~Khan,
        Fahad~Shahbaz~Khan, 
        Ling~Shao,
        and~Ming-Hsuan~Yang
        
\IEEEcompsocitemizethanks{\IEEEcompsocthanksitem J. Rajasegaran,  M. Hayat, S. Khan, F. S. Khan and L. Shao are with the Inception Institute of Artificial Intelligence, Abu Dhabi, UAE.\protect\\
E-mail: jathushan.rajasegaran@inceptioniai.org
}
\IEEEcompsocitemizethanks{\IEEEcompsocthanksitem Ming-Hsuan~Yang is with University of California at Merced, USA.\protect\\

}
\thanks{Manuscript received April 19, 20--; revised August 26, 20--.}}

\markboth{Journal of \LaTeX\ Class Files,~Vol.~14, No.~8, August~2015}%
{Rajasegaran \MakeLowercase{\textit{et al.}}: Random Path Selection for Incremental Learning}

\IEEEtitleabstractindextext{%
\begin{abstract}
In a conventional supervised learning setting, a machine learning model has access to examples of `all' object classes that are desired to be recognized during the inference stage. This results in a fixed model that lacks the flexibility to adapt to new learning tasks. In practical settings, learning tasks often arrive in a sequence and the models must continually learn to increment their previously acquired knowledge. Existing incremental learning approaches fall well below the state-of-the-art cumulative models that use all training classes at once. In this paper, we propose a random path selection algorithm, called \ours{}, that progressively chooses optimal paths for the new tasks while encouraging parameter sharing between tasks. We {introduce} a new network capacity measure that enables us to automatically switch paths if the already used resources are saturated. Since the proposed {path-reuse} strategy ensures forward knowledge transfer, our approach is efficient and has considerably less computation overhead. As an added novelty, the proposed model integrates knowledge distillation and retrospection along with the path selection strategy to overcome catastrophic forgetting. In order to maintain an equilibrium between previous and newly acquired knowledge, we propose a simple controller to dynamically balance the model plasticity.  Through extensive experiments, we demonstrate that the \ours{} method surpasses the state-of-the-art performance {for} incremental learning and by utilizing parallel computation this method can run in constant time with nearly the same efficiency as a conventional deep convolutional neural network. Our codes are available at: \url{https://github.com/brjathu/RPSnet}.

\end{abstract}

}

\maketitle

\IEEEdisplaynontitleabstractindextext

\IEEEpeerreviewmaketitle
\section{Introduction}

In several real-life applications, it is desired to continually update a model as new object classes are encountered. Conventional deep neural networks work in a \emph{static} setting, where a model is trained on the training samples from \emph{all} classes of interest. Such models can only recognize the set of classes initially used for training, thereby lacking the ability to incrementally learn new tasks. As such, when these networks are sequentially trained on a series of tasks, they suffer from `\emph{catastrophic forgetting}' \cite{mccloskey1989catastrophic} thereby degrading the performance on old tasks. A typical example is the transfer learning scenario where a model pre-trained on a source task is fine-tuned to a target task, resulting in overriding the previously learned information and consequently degrading the performance on source task \cite{khan2018guide}. Therefore, it is fundamentally important to develop models capable of incrementally adding new classes without the need to retrain models from scratch using all previous class-sets.


A versatile incremental learning model possess the following characteristics. \textbf{(a)} As a model learns new tasks, its performance on old tasks should not catastrophically deteriorate (the forgetting problem). \textbf{(b)} Since learning tasks are inherently related, the knowledge acquired on old tasks should help in accelerating the learning on new tasks (i.e., forward transfer). \textbf{(c)} Ensure that complimentary representations are learned from the current task so that the newly learned information can help improve the old task performance (i.e., backward transfer). \textbf{(d)} As the class-incremental learning progresses, the network must share and reuse the previously tuned parameters to ensure a bounded computational complexity and memory footprint of the final model. \textbf{(e)} At all learning stages, the model must maintain a tight equilibrium between the existing knowledge base and newly presented information, thereby addressing stability-plasticity dilemma. 



Addressing all the above requirements is a challenging task, making incremental learning an open research problem.  Among seminal previous works, \cite{li2018learning} employs a distillation loss to preserve knowledge across multiple tasks but requires prior knowledge about the task corresponding to a test sample during inference. An incremental classifier and representation learning approach \cite{rebuffi2017icarl} jointly uses distillation and prototype rehearsal but retrains the complete network for new tasks, thus reducing model's stability. The progressive network \cite{rusu2016progressive} grows paths linearly as new tasks arrive that in turn reduces scalability since the parameters rise quadratically in proportion to the number of tasks. The elastic weight consolidation scheme \cite{kirkpatrick2017overcoming} computes synaptic importance of weights that helps in overwriting less important parameters. However, the scalability is still an issue since the Fisher information metric used to estimate synaptic importance is computed offline and while the approach works quite well for permutation tasks, its performance suffers on class-incremental learning \cite{kemker2018measuring}. 


This work is based on the idea that the most important characteristic of a true incremental learner is to maintain the right trade-off between `\emph{stability}' (leading to \emph{intransigence}) and `\emph{plasticity}' (resulting in \emph{forgetting}). We achieve this requisite via a dynamic path selection approach, called \ours{}, that automatically estimates the network capacity and selects a new trainable path to learn complimentary features, when required. Our approach proceeds with random candidate paths and discovers the optimal one for a given task. Once a task is learned, we fix the parameters associated with it, that can only be shared by future tasks. To enable parameter reuse and minimize model complexity, \ours{} utilizes the same path until the network saturates. To complement the previously learned representations, we propose a stacked residual design that focuses on learning the supplementary features suitable for new tasks. Besides, our learning scheme 
introduces an explicit controller module to maintain the equilibrium between stability and plasticity for all tasks. During training, our approach always operates with a constant parameter budget that at max equals to a conventional linear model (e.g., \texttt{resent} \cite{he2016deep}). Furthermore, it can be straightforwardly parallelized during both train and test stages. With these novelties, our approach performs favorably well on the class-incremental learning results, surpassing the previous best model~\cite{rebuffi2017icarl} by $7.38\%$ and $10.64\%$ on CIFAR-100 and ImageNet datasets, respectively.

Our main contributions are:\vspace{-6pt}
\begin{itemize}\setlength{\itemsep}{0em}
    \item A random path selection approach that enables forward and backward knowledge transfer via efficient path sharing and reuse. Importantly, using a new path saturation measure, it automatically switches paths when network's capacity is fully utilized. 
    \item The residual learning framework that incrementally learns residual paths which allows network reuse and accelerate the learning process resulting in faster training as new tasks arrive.
    \item Ours is a hybrid approach that combines the respective strengths of knowledge distillation (via regularization), retrospection (via exemplar replay) and dynamic architecture selection methodologies to deliver a strong incremental learning performance.
    \item A novel controller that guides the plasticity of the network to maintain an equilibrium between the previously learned knowledge and the new tasks. 
\end{itemize}

\begin{figure}[t]
    \centering
    \includegraphics[scale=1.4, angle=-90]{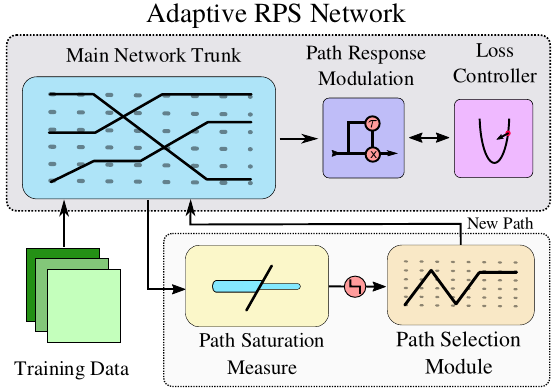}
    \caption{\emph{\ours{} Architecture.} Our incremental learning approach reuses the previous knowledge for the upcoming tasks without sacrificing the performance on old tasks. This is achieved via a novel path selection algorithm and a loss controller that avoids catastrophic forgetting. To maintain a low computational overhead, \ours{} selects new paths by measuring the saturation level of the network, thereby ensuring efficiency and good performance.}
    \label{fig:new_arch}
\end{figure}

A preliminary version of our approach was published at the NeurIPS conference in 2019 \cite{rajasegaran2019random}. In the current work, we considerably improve \cite{rajasegaran2019random} by introducing: (a) an automatic path switching mechanism based on a novel network capacity measure (Sec.~\ref{sec:network_capacity}), (b) a path attention approach to appropriately modulate the response from individual network paths (Sec.~\ref{sec:peak_path_response}), (c) additional experiments on the large-scale MS-Celeb and ImageNet-1K datasets, and (d) new comparative and ablation analysis e.g., with different switching rules, with a Genetic algorithm and for various attention functions (Sec.~\ref{sec:ablation}). Overall, our new  \ours{} framework delivers better performance and is computationally more efficient compared to the conference version~\cite{rajasegaran2019random}. Fig.~\ref{fig:new_arch} shows the proposed Adaptive RPS-Net architecture.


\section{Related Work}
The catastrophic interference problem was first noted to hinder the learning of connectionist networks by  \cite{mccloskey1989catastrophic}. This highlights the stability-plasticity dilemma in neural networks \cite{abraham2005memory} i.e., a rigid and stable model will not be able to learn new concepts while an easily adaptable model is susceptible to forget old concepts due to major parameter changes. The existing continual learning schemes can be divided into a broad set of three categories: \textbf{(a)} regularization schemes, \textbf{(b)} memory based retrospection and replay, and \textbf{(c)} dynamic sub-network training and expansion.

\emph{\textbf{Regularization based Methods:}} A major trend in continual learning research has been on proposing novel regularization schemes to avoid catastrophic forgetting by controlling the plasticity of network weights. \cite{li2018learning} proposed a knowledge distillation loss \cite{hinton2015distilling} which forces the network to retain its predictions on the old tasks. The distillation loss was also used for class-incremental setting in \cite{rebuffi2017icarl,Castro_2018_ECCV} to minimize forgetting. Kirkpatrick et al.~\cite{kirkpatrick2017overcoming} proposed an elastic weight consolidation mechanism that quantifies the relevance of parameters to a particular task and correspondingly adjusts the learning rate. In a similar spirit,  \cite{zenke2017continual} designed intelligent synapses which measure their relevance to a particular task and consequently adjust plasticity during learning to minimize interference with old tasks. Compared to these works, we also use a regularization loss but more importantly, we automatically estimate the network capacity using a new path saturation measure that helps us fix a network module upon saturation.

\emph{\textbf{Memory based Approaches:}} Rebuffi et al. \cite{rebuffi2017icarl} proposed a distillation scheme intertwined with exemplar-based retrospection to retain the previously learned concepts. \cite{hou2018lifelong} considered a similar approach for cross-dataset continual learning \cite{li2018learning}. The combination of episodic (short-term) and semantic (long-term) memory  was studied in \cite{kemker2017fearnet,gepperth2016bio,kamra2017deep} to perform memory consolidation and retrieval. Particularly, \cite{kamra2017deep,kemker2017fearnet} help avoid explicitly storing exemplars in the memory, rather using a generative process to recall memories. Along similar lines, \cite{xiang2019incremental} proposed an adversarial generative model to create pseudo-examples of the old tasks.  Recently, \cite{belouadah2019il2m} proposed a dual-memory approach that not only stores the old task exemplars but also the class-specific statistics from the originally learned model. Different from these works that offer less scalability when number of tasks increase, we maintain a fixed memory budget to keep task exemplars and fix parts of the network to retain old information.

\emph{\textbf{Dynamic Architectures:}} The third stream of works explores dynamically adapting network architectures to cope with the growing learning tasks.  \cite{rusu2016progressive} proposed a network architecture that  progressively adds new branches for novel tasks that are laterally connected to the fixed existing branches. Similarly,  \cite{xiao2014error} proposed a network that not only grows incrementally but also expands hierarchically by grouping together the similar classes. 
Specific paths through the network were selected for each learning task using a genetic algorithm in PathNet \cite{fernando2017pathnet}. Afterwards, task-relevant paths were fixed and reused for new tasks to speed-up the learning efficiency. 

The existing adaptive network architectures come with their respective limitations e.g., \cite{rusu2016progressive}'s complexity grows with the tasks, \cite{xiao2014error} has an expensive training procedure and a somewhat rigid architecture 
and \cite{fernando2017pathnet} does not allow incrementally learning new classes due to a detached output layer and the genetic learning algorithm used in~\cite{fernando2017pathnet} is relatively inefficient. In comparison, we propose a random path selection methodology that provides a performance boost alongwith a faster convergence. Furthermore, our approach combines the respective strengths of the above two types of methods by introducing a distillation procedure alongside an exemplar-based memory replay to avoid catastrophic forgetting. 

\emph{\textbf{Differences with PathNet:}} The closest to our approach is PathNet \cite{fernando2017pathnet}, yet there are notable differences in terms of network architecture, learning methodology and the end problem. In terms of architecture, PathNet adds all modules in a simple feed-forward manner, while RPS-Net adds modules as residual connections to an identity path 
which can always be learnt even when all the modules are saturated. RPS-Net can therefore adapt to new changes more robustly. During training, PathNet uses an evolutionary algorithm while RPS-Net employs a random path selection methodology. Since PathNet usually involves only two tasks, an evolutionary algorithm with large explorations/generations is viable. However, for our case i.e., $10+$ tasks, PathNet is not feasible, due to a large number of possible explorations. Furthermore, our proposed architecture is more suitable for incremental learning and offers several novel components such as a dynamic path switching, a controller to maintain network's elasticity-plasticity balance, and a goal driven path attention methodology.

\begin{figure*}[!ht]
    \centering
    \includegraphics[scale=0.12,angle=-90]{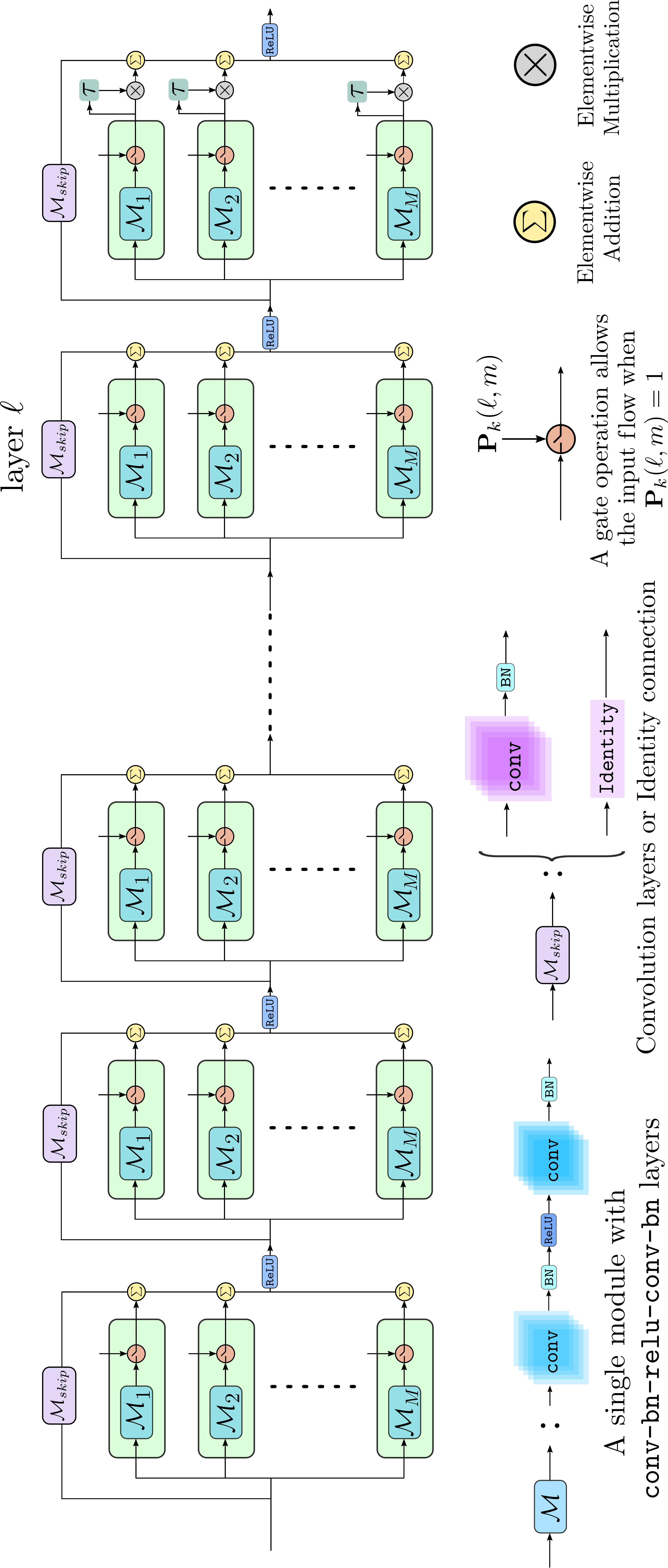}
    \caption{\emph{An overview of our \ours{}:} The network architecture utilizes a parallel residual design where the optimal path is selected among a set of randomly sampled candidate paths for new tasks. The residual design allows forward knowledge transfer and faster convergence for later tasks.  An attention mechanism ($\tau$) is applied on the module outputs from the final layer to appropriately re-weight contributions from previously learned tasks. Our approach is trained with a hybrid objective function that ensures the right trade-off between network stability and plasticity, thus avoiding catastrophic forgetting.}
    \label{fig:network}
\end{figure*}

\section{Our Approach}

We consider the recognition problem in an incremental setting where new tasks are sequentially added. Assume a total of $K$ tasks, each comprising of $U$ classes. Our goal is to sequentially learn a deep neural network, that not only performs well on the new tasks but also retains its performance on the old tasks. To address this problem, we propose a random path selection approach (\ours{}) that progressively builds on the previously acquired knowledge to facilitate faster convergence and better performance. In the following, we explain the novel aspects of \ours{} that include the proposed network architecture (Section~\ref{sec:arch}), our path selection strategy (Section~\ref{sec:network_capacity} and \ref{sec:path_sel}) and a hybrid objective function (Section~\ref{sec:obj}). 

\subsection{\ours{} Architecture}\label{sec:arch}
Our network consists of $L$ distinct layers (see Figure \ref{fig:network}). Each layer $\ell \in [1,L]$ constitutes a set of basic building blocks, called modules $\mathcal{M}^{\ell}$. For simplicity, we consider each layer to contain an equal number of $M$ modules, stacked in parallel, i.e., $\mathcal{M}^{\ell} = \{\mathcal{M}^{\ell}_m\}_{m=1}^{M}$, along with a skip connection module $\mathcal{M}_{skip}^{\ell}$ that carries the bypass signal. The skip connection module $\mathcal{M}_{skip}^{\ell}$ is an identity function when the feature dimensions do not change and a learnable module when the dimensions vary between consecutive layers. A module $\mathcal{M}^{\ell}_m$ is a learnable sub-network that maps the input features to the outputs. In our case, we consider a simple combination of  \texttt{(conv-bn-relu-conv-bn)} layers for each module, similar to a single \texttt{resnet} block \cite{he2016deep}. In contrast to a residual block which consists of a single identity connection and a residual branch, we have one skip connection and $M$ residual blocks stacked in parallel. The intuition behind developing such a parallel architecture is to ensure multiple tasks can be continually learned without causing catastrophic interference with other paths, while simultaneously providing parallelism to ensure efficiency. 

Towards the end of each layer in our network, all the residual connections, as well as skip connections, are  combined together using element-wise addition to aggregate complimentary task-specific features obtained from different paths. Notably, for the base-case when $M=1$, the network is identical to a conventional \texttt{resnet} model. After the Global Average Pooling (GAP) layer that collapses the input feature maps to generate a final feature $\mathbf{f}\in \mathbb{R}^{D}$, we use a fully connected layer classifier with weights $\mathbf{W}_{fc} \in \mathbb{R}^{D\times C}$ ($C$ being the total number of classes) that is shared among all tasks. As such, we have $C = K \times U$.

\subsubsection{Training and Inference Paths}
For a given \ours{}, with $M$ modules and $L$ layers, we can define a path $\mathbf{P}_k \in \mathbb{R}^{L \times M}$ for a task $k$, such that:
\begin{align}
    \mathbf{P}_k (\ell,m) = 
    \begin{cases} 
    1, & \text{if the module $\mathcal{M}^{\ell}_m$ is added to the path,} \\
    0, & \text{otherwise}.
    \end{cases}
\end{align}
The path $\mathbf{P}_k$ is basically arranged as a stack of one-hot encoded row vectors $\bm{e}^{(i)}$ (with $i^{th}$ standard basis): 
\begin{align*}\label{eq:rand_path}
\mathbf{P}_k = \Big\{ \mathbf{P}_k(\ell) \in \{0,1\}^{M} : \mathbf{P}_k(\ell) &= \bm{e}^{(i)} \equiv \sum_{m=1}^{M} \mathbf{P}_k (\ell,m) = 1\Big\}, \\& s.t., i \sim \mathbb{U}\big(\{\mathbb{Z} \cap [1,M]\}\big),
\end{align*}
where $i$ is the selected module index, uniformly sampled using $\mathbb{U}(\cdot)$ over the set of integers $[1,M]$.

We define two set of paths $\mathbf{P}^{tr}_k$ and $\mathbf{P}^{ts}_k$ that denote the train and inference paths, respectively. Both are formulated as binary matrices: $\mathbf{P}^{tr,ts}_k \in \{0,1\}^{L\times M}$. When training the network, any $m^{th}$ module in $\ell^{th}$ layer with $\mathbf{P}^{tr}_k(\ell,m)=1$ is activated and all such modules together constitute a training path $\mathbf{P}^{tr}_k$ for task $k$. 
As we will elaborate in Section~\ref{sec:path_sel}, the inference path is evolved during training by sequentially adding newly discovered training paths and ends up in a ``\emph{common}'' inference path for all inputs, therefore our \ours{} does not require knowledge about the task an input belongs to. Some popular methods (e.g., \cite{li2018learning,lopez2017gradient}) need such information, which limits their applicability to real-world incremental class-learning settings where one does not know in advance the corresponding task for an input sample.  Similar to training, only the modules with $\mathbf{P}^{ts}_k(\ell, m)=1$ are used in the inference stage.

\subsubsection{Attention based on Peak Path Response}\label{sec:peak_path_response}
In \ours{}, the output tensors from each module in layer $\ell$ are jointly used to compute the input to the successive layer. A naive way to aggregate feature tensors $\mathbf{T}_{\ell,m}$  is to obtain an accumulative response by assuming an equal contribution from all paths in layer $\ell$:
\begin{equation}
    \label{eq:next_tensor}
    \mathbf{T'}_{\ell} = \sum_{m=1}^{M} \mathbf{P}_k (\ell,m)  \cdot \mathbf{T}_{\ell,m} ,
\end{equation}
where, $\mathbf{T}_{\ell,m}$ is the output after the convolution from a module $m$ in layer $\ell$ and $\mathbf{T'}_{\ell}$ denotes the accumulated feature response forwarded to the next layer. This raises an important question: \emph{Do all previous paths (corresponding to old tasks) contribute equally to learn a new task?} For example, for a new task with images of \texttt{birds}, a path previously trained with \texttt{car} images may contribute less than a path trained with \texttt{animals}. Therefore, we propose a simple Peak Path Response (PPR) based attention mechanism that is used to appropriately re-weight the contributions from different paths.  In simple words, the peak response for a path signifies its relevance to the given input belonging to a new task. Specifically, PPR attention multiplies each convolution layer output tensor by its corresponding path response value $\tau(\mathbf{T}_{\ell,m} )$ as shown below: 
\begin{equation}
    \label{eq:next_tensor_attention}
    \mathbf{T'}_{\ell} = \sum_{m=1}^{M} \tau(\mathbf{T}_{\ell,m} ) \cdot \mathbf{P}_k (\ell,m)  \cdot \mathbf{T}_{\ell,m} .
\end{equation}
Here, $\tau$ is a scalar value function which maps $ h \times w \times c$ tensor to a scalar weighting factor. We use $\tau(\mathbf{T}_{\ell,m} ) = \max(\mathbf{T}_{\ell,m})$ as the PPR coefficient for each path. We also tried several other mapping functions, such as 1) path mean response, 2) computing the maximum activation along the feature maps and converting $c$ feature maps to a single scalar with a fully connected layer, and 3) using $1\times1$ convolution to map $c$ features into single $h\times w$ map and then passing it through a fully connected layer for a scalar coefficient. However, the best performance is achieved with the peak response coefficient as we illustrate via the ablation study reported in Section~\ref{sec:ablation}. Our experiments show that the proposed PPR attention delivers best performance when used in the final layer $L$ with $\mathbf{T}_{L,m} \in \mathbb{R}^{8 \times 8 \times 512}$ just before classification. This is intuitive since a re-weighting should be applied only when high-level representations are extracted that encapsulate class-semantics (i.e., close to final layers). Furthermore, the attention operation in the final layer helps in achieving our desired behaviour i.e., re-weighting the overall response of each path based on its relevance to a new task.

\subsection{Measuring Network Capacity}\label{sec:network_capacity}

As we progress by adding more tasks in an  incremental learning setting, it is likely that the model saturates. This prevents the deep network to adapt to new tasks. Our proposed network architecture is designed to solve this problem by switching new paths as network starts to saturate. A naive solution would be choosing new paths for a fixed set of new tasks. For example, one can use a single path to train for every two consecutive tasks. 
Although this solves the network saturation problem, having a \emph{fixed rule} for changing paths (e.g., after every $n$ tasks) is a sub-optimal solution. We suggest that the path switching mechanism should be based on the network capacity that will ensure a dynamic and efficient resource expansion/reuse. For example, generally earlier tasks are more divergent and require quick path switching, while later tasks can reuse the early learned features, hence several tasks can use a single path in the later training stages. Since, measuring network capacity is a non-trivial task, we propose a new measure for network saturation that helps us decide the point when paths must be switched. 

Our saturation measure is based on the Fisher information matrix, that estimates the second-order derivatives of the loss function close to a minima. Given a likelihood function $f(\theta)$ for parameters $\theta$ conditioned on input data distribution $\mathcal{X}$, the Fisher information matrix $\mathbf{F}$ can be represented as:
\begin{align}
\mathbf{F} = \underset{f(\mathcal{X}; \theta)}{\mathbb{E}}\big[ \nabla \log f(\mathcal{X}; \theta) \nabla \log f(\mathcal{X}; \theta)^T  \big].
\end{align} 
The matrix $\mathbf{F}$ is basically the expectation of Hessian but can be approximated using only first-order derivatives and is positive semi-definite in all cases. 
The diagonal Fisher information or the precision (inverse of variance) tells the significance of a specific parameter for the correct classification of task $k$. In other words, if the variance of the parameter is high or precision is low, it shows that the corresponding parameter has less contribution in the final classification and vice versa. 

Given the Fisher information, we define a measure for network saturation. First, we compute the maximum values of diagonal Fisher information matrix over all the exemplar set images. Notably, this is in contrast to how Fisher information has been used previously in the literature, e.g., \cite{kirkpatrick2017overcoming} used the `\emph{average}'  diagonal Fisher information for all samples in an exemplar set to define the elasticity of parameters for an old task. In our case, averaging out the Fisher information is counter-intuitive because if a parameter $\theta_a$  is significant for task $k$ and has no significant contribution for any other task, then an average over all the exemplars (which contains samples from task $k$ as well as old tasks) would undermine the parameter $\theta_a$'s significance. Therefore, $\mathbf{I}$ stores the element-wise maximum values of Fisher information matrices $\{\mathbf{F}_i\}$ for all images in the exemplar set,
\begin{equation}\label{eq:exemplar_fisher}
    \mathbf{I}^{ij} = \max_{} \big( { \mathbf{F}^{ij}_1, \mathbf{F}^{ij}_2, \ldots \mathbf{F}^{ij}_{|\mathcal{E}|} } \big).
\end{equation}
 Here, $\mathcal{E}$ contains exemplar images from all previous tasks and $|\cdot|$ denotes set cardinality. Keeping the maximum values from the 
 Fisher matrix helps to keep significant parameters which capture independent information irrespective of the task.

Further, as we progress with learning new tasks, the magnitude of $\mathbf{I}$ entries keep on decreasing. This is because the $\mathbf{I}$ is approximated with first-order gradients, as the calculation of second-order gradients on a deep convolutional network is computationally very expensive. Therefore, we calculate the `\emph{relative significance}' between the sets of early and final layers. If a high percentage of parameters are important for previous tasks then the network is likely to be saturated. Hence, we define relative saturation coefficient $\mu_{sat}$ as,
\begin{equation}
    \mu_{sat} = \log \bigg( \frac{\frac{1}{|fl|} \sum_{\ell \in fl} \mathrm{tr}(\mathbf{I}_{\ell} )}{ \frac{1}{|el|} \sum_{m \in el}   \mathrm{tr}(\mathbf{I}_{m}) }   \bigg).
\end{equation}
where $\mathrm{tr}$ denotes trace of the matrix, $\mathbf{I}_{\ell}$ is the Fisher information matrix for $\ell^{th}$ layer and $fl$ and $el$ are sets of layers at the end and start of the trainable path of the network, respectively. Note that $\mu_{sat}$ is calculated only with the images in the exemplar set (see Equation~\ref{eq:exemplar_fisher}). We use ten layers including kernel weights and biases in each set. Having a set of layers for this calculation helps suppress the noise introduced in Fisher information matrix by individual weights.

\ours{} architecture is designed in such a way that a new path will learn the information not captured by the previously learned paths (since any path can be considered as a residual connection for the rest). Therefore, it is reasonable to assume that parameters in different paths will learn complimentary (and thus independent) information. This is why path switching is performed in our framework only when the trainable path is saturated for a given task (meaning all parameters have been fully utilized for previous task's learning) and a substitute path is required to carry on learning new tasks. This way, although all tasks are not fully independent, the features learned by each path are independent. Moreover, for a given trainable path, parameters are tuned using exemplars from previous tasks and training samples of the current task. Therefore, the newly learned parameters for the current task focus on learning features that have not been encapsulated during the old tasks learning.

\begin{figure*}[!t]
    \centering
    \includegraphics[scale=0.037,angle=-90]{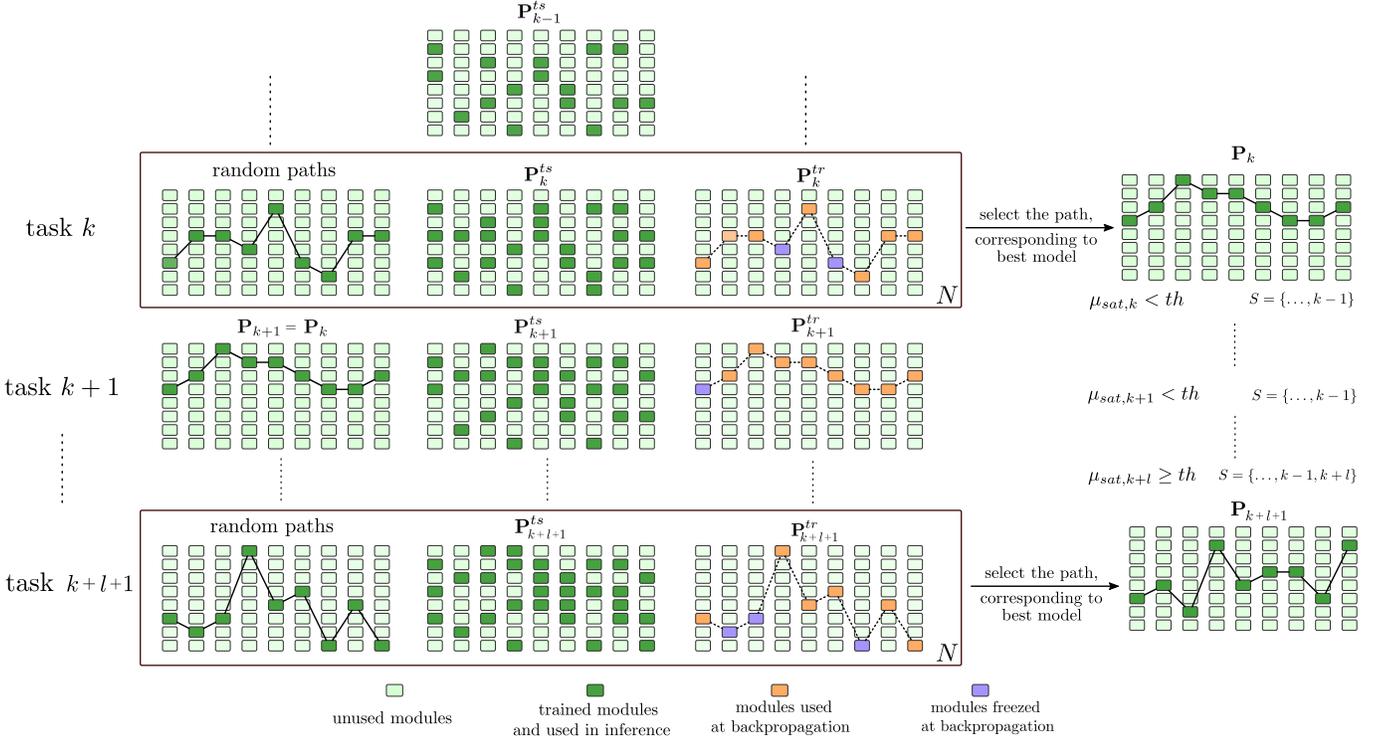}
    \caption{\emph{Path Selection Approach:} Given a task $k$, $N$ random paths are initialized. For each path, only the modules different from the previous inference path $\mathbf{P}_{k-1}^{ts}$ are used to form the training path $\mathbf{P}_{k}^{tr}$. Among $N$ such paths, the optimal $\mathbf{P}_k$ is selected  and combined with the $\mathbf{P}_{k-1}^{ts}$ to obtain $\mathbf{P}_{k}^{ts}$. Notably, the path selection is only performed if $\mu_{sat,k} \geq th$. In above scenario, $\mu_{sat,k}, \mu_{sat,k+1}, \dots,  \mu_{sat,k+l-1}$ are below the threshold $th$. Hence, the same path is used for training tasks $k, k+1, \dots , k+l$, and $k+l$ is added to the list $S$, such that $S=\{\dots, k+l\}$ and a new path is selected for the next task $k+l+1$. During training, the complexity remains bounded by a standard single path network and the resources are shared between tasks. }
    \label{fig:path_sel}
\end{figure*}

\subsection{Path Selection Approach}\label{sec:path_sel}
\ours{} is a progressive learning approach. The path selection scheme enables incremental and bounded resource allocation that ensures knowledge exchange between the old and new tasks resulting in positive forward and backward transfer.
As discussed in the previous section, we use path saturation coefficient as a simple measure to decide whether we need to switch the current path or not. For a given threshold value `$th$', after training for a task, we simply calculate the saturation coefficient in the current path configuration. If the network saturation is above the threshold $th$, a set of $N$ new  paths are randomly sampled. At this point, we stop the training of the old modules (i.e., fix their paths and parameters). We then train $N$ models in parallel for the next task and chose the best one, otherwise we keep training the network with the same path configuration as before. Hence, at any point, only $L$ layers (each with a maximum of one module) are being trained.


Path selection strategy balances the trade off between resource usage (number of neurons) and performance. If we set a small value for $th$, for a given number of total tasks, the network will choose more paths to keep the saturation low. On the other hand, with a high threshold value, the network will reuse a certain path many times until it reaches the specified $th$ value. Therefore, with small thresholds the network takes more resources in terms of parameters, training and inference time, while for a higher $th$ value, network will use comparatively less resources but at the cost of a slight drop in performance (see Section~\ref{sec:ablation} for experimental results). Nevertheless, we noted that an adaptive path switching rule based on path saturation almost always perform better than a fixed rule (path switching after every $n$ tasks) with comparable resource usage. 


The path selection strategy based on network saturation is illustrated in Figure~\ref{fig:path_sel}. Our choice of random path generation as a mechanism to select an optimal path is mainly inspired by the recent works of \cite{xie2019exploring,zoph2018learning,pham2018efficient}. These works show that random search for an optimal network architecture performs somewhat comparable to other computationally demanding approaches e.g., 
the ones based on reinforcement learning (RL). Besides, some incremental learning approaches add new resources to the network, resulting in network expansion \cite{rusu2016progressive, xiao2014error}.
In contrast, our path selection algorithm does not result in linear expansion of resources since a new path is created only if the current path saturates and overlapping modules are reused when the new path is intersecting with old paths. Further, even when all the modules are exhausted (saturated), the skip connections are always trained. We show via an extensive ablation study (Section~\ref{sec:ablation}) that even when all paths are saturated, our \ours{} can still learn useful representations as the skip connections and the final classification layer remains tunable in every case.


At any point in time, we train a single path (equivalent to a \texttt{resnet}) while rest of the previously trained paths are fixed. Due to this, the path we use for a task $k$ essentially learns the residual signal relative to the fixed paths that were previously trained for old tasks. For example, if we are training $\textbf{P}^{tr}_k$, the weights of $\mathbf{P}^{ts}_{\lfloor S_{-1} \rfloor} \veebar \textbf{P}^{tr}_k$  are fixed, where $\veebar$ denotes the exclusive disjunction (logical XOR operation) and $S$ is a list containing the tasks numbers where network saturates above threshold value (resulting in path switching). $S_{-1}$ returns the last task number where saturation along the training path exceeds $th$ and $k=S_{-1} + 1$ will be trained with a new random path. Essentially, the complete $\textbf{P}^{tr}_k$ is not used for training rather its disjoint portion that has not already been trained for any of the old tasks is learned i.e., $\textbf{P}^{tr}_k \veebar (\textbf{P}^{tr}_k \land \textbf{P}^{ts}_{\lfloor S_{-1} \rfloor})$, where $\land$ denotes logical conjunction (logical and) operator. In this way,  previous knowledge is shared across the network via overlapping paths and skip connections. When the network is already trained for several tasks, a new path for the current task only needs to learn higher order residual signal in the network. This has an added advantage that convergence becomes faster as we learn more tasks since each new task will be learned taking advantage of the previous information. 

The optimal path based on the performance of $N$ path configurations is selected as $\mathbf{P}_k$. All such task-specific paths are progressively combined together to evolve a common inference path $\textbf{P}^{ts}_k$,
\begin{equation}
    \textbf{P}^{ts}_k =  \mathbf{P}^{tr}_1 \lor \mathbf{P}^{tr}_2  \ldots \lor \mathbf{P}^{tr}_k,
\end{equation}
where $\lor$ denotes the inclusive disjunction (logical OR) operation. At each task $k$, the inference path $\textbf{P}^{ts}_k$ is used to evaluate all previous classes.

\subsection{Incremental Learning Objective}\label{sec:obj}

\textbf{Loss function:} We use a hybrid loss function that combines regular cross-entropy loss as well as a distillation loss to incrementally train the network.

 For a task $k\in[1,K]$ which contains $U$ classes, we calculate the cross-entropy loss as follows,
\begin{equation}
    \mathcal{L}_{ce} = - \frac{1}{n}\sum\limits_{i} {\bf t}_i[1:k * U] \log(\text{softmax}({\bf q}_i[1:k * U]))
\end{equation}
where $i$ denotes the example index, ${\bf t}(x)$ is the one-hot encoded true label, ${\bf q}(x)$ are the logits obtained from the network's last layer and $n$ is the mini batch size. To keep the network robust to catastrophic forgetting, we also use a distillation loss in the objective function,
\begin{align*}
    \mathcal{L}_{dist} =  \frac{1}{n} \sum\limits_{i} \text{KL}\bigg(  \log\bigg(\sigma&\bigg(\frac{{\bf q}_i[1:(k-1) * U]}{t_e}\bigg)\bigg), \\&\sigma\bigg(\frac{{\bf q'}_i[1:(k-1) * U]}{t_e}\bigg)   \bigg). 
\end{align*}
Here, $\sigma$ is the softmax function, $t_e$ is the temperature used in~\cite{hinton2015distilling} and ${\bf q'}(x)$ are the logits obtained from the network's previous state.

\textbf{Controller:} It is important to maintain a  balance between the previously acquired learning and the knowledge available from the newly presented task. If the learning is biased towards either of the two objectives, it will result in either catastrophic forgetting (losing old task learning) or interference (obstructing learning for the new task). Since our network is trained with a combined objective function with $\mathcal{L}_{ce}$ and $\mathcal{L}_{dist}$, it is necessary to adequately control the plasticity of the network. We propose the following controller that seeks to maintain an equilibrium between $\mathcal{L}_{ce}$ and $\mathcal{L}_{dist}$,
\begin{equation}
    \mathcal{L} = \mathcal{L}_{ce} + \phi(k, \gamma) \cdot \mathcal{L}_{dist}.
\end{equation}
Here, $\phi(k, \gamma)$ is a scalar coefficient function, with $\gamma$ as a scaling factor, introduced to increase the distillation contribution to the total loss. Intuitively, as we progress through training, $\phi(k, \gamma)$ will also increase to ensure that network remembers old information. It is defined as,
\begin{equation}
    \phi(k, \gamma) = \begin{cases}1, \ \ \ \ \ \ \ \ \ \ \ \ \ \ \ \  \ \ \ \ \text{ if $k \leq S_0$} \\ 
    (k-S_0) * \gamma, \ \ \ \text{ otherwise}.\end{cases}
\end{equation}
Here, $S_0$ is the task number where the network changes the path configuration for the first time. 

\section{Experiments and Results}
\subsection{Implementation Details}
\textbf{Dataset and Protocol:} For our experiments, we use evaluation protocols similar to iCARL \cite{rebuffi2017icarl}. We incrementally learn $100$ classes on CIFAR-100 in groups of $10$, $20$ and $50$ at a time. For ImageNet, we use the same subset as \cite{rebuffi2017icarl} comprising of $100$ classes and incrementally learn them in groups of $10$. After training on a new group of classes, we evaluate the trained model on test samples of all seen classes (including current and previous tasks). Following iCARL \cite{rebuffi2017icarl}, we restrict exemplar memory budget to $2$k samples for CIFAR-100 and 
ImageNet datasets. Note that unlike iCARL, we randomly select our exemplars and do not employ any herding and exemplar selection mechanism.

We also experiment our model with MNIST and SVHN datasets. For this, we resize all images to $32 {\times} 32$ and keep a random exemplar set of $4.4$k, as in \cite{Hsu18_EvalCL}. We group $2$ consecutive classes into one task and incrementally learn  five tasks. For evaluation, we report the average over all classes ($A_5$).  

\begin{figure*}[htp]
\center
\includegraphics[width=0.45\textwidth]{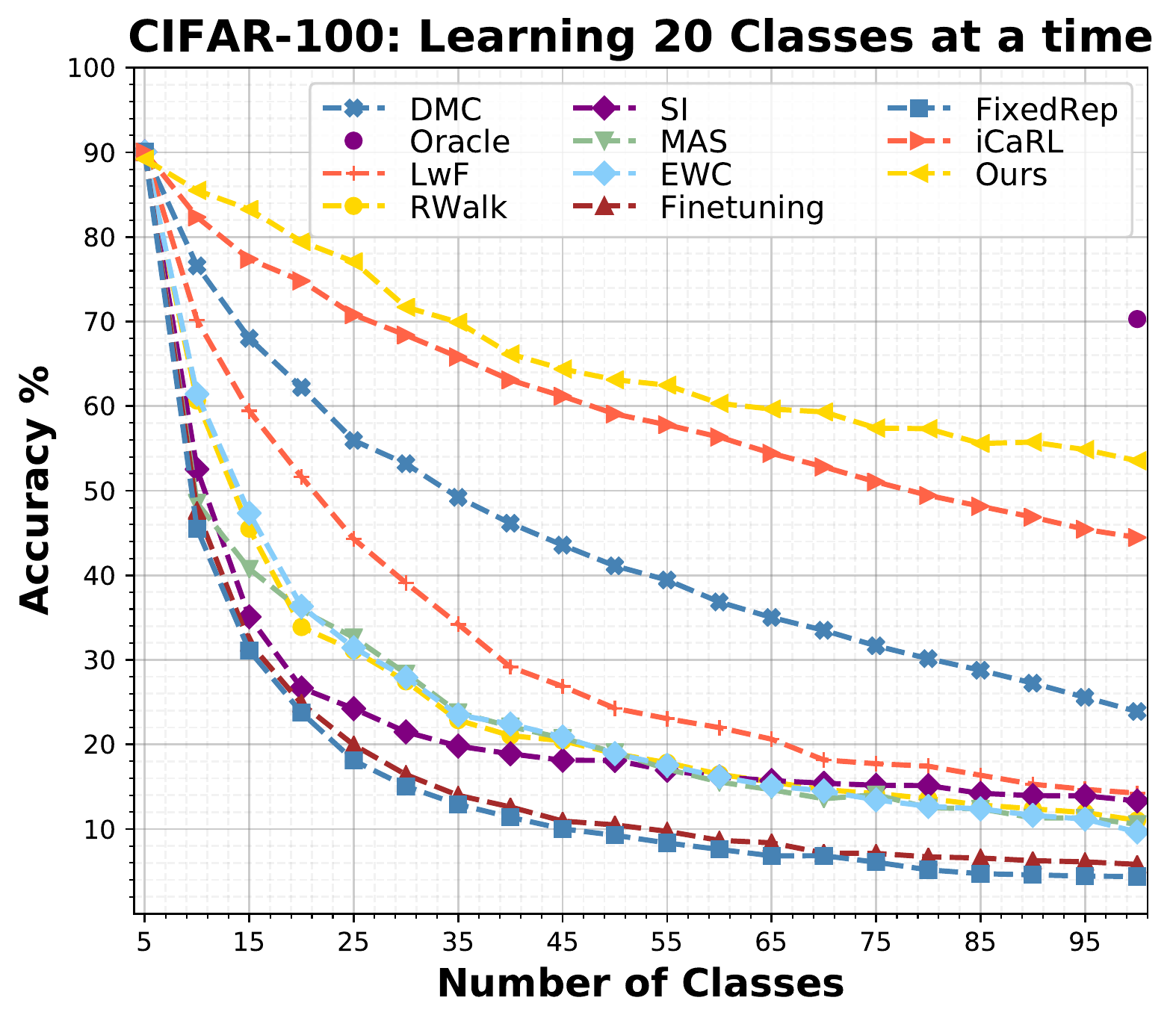}
\includegraphics[width=0.45\textwidth]{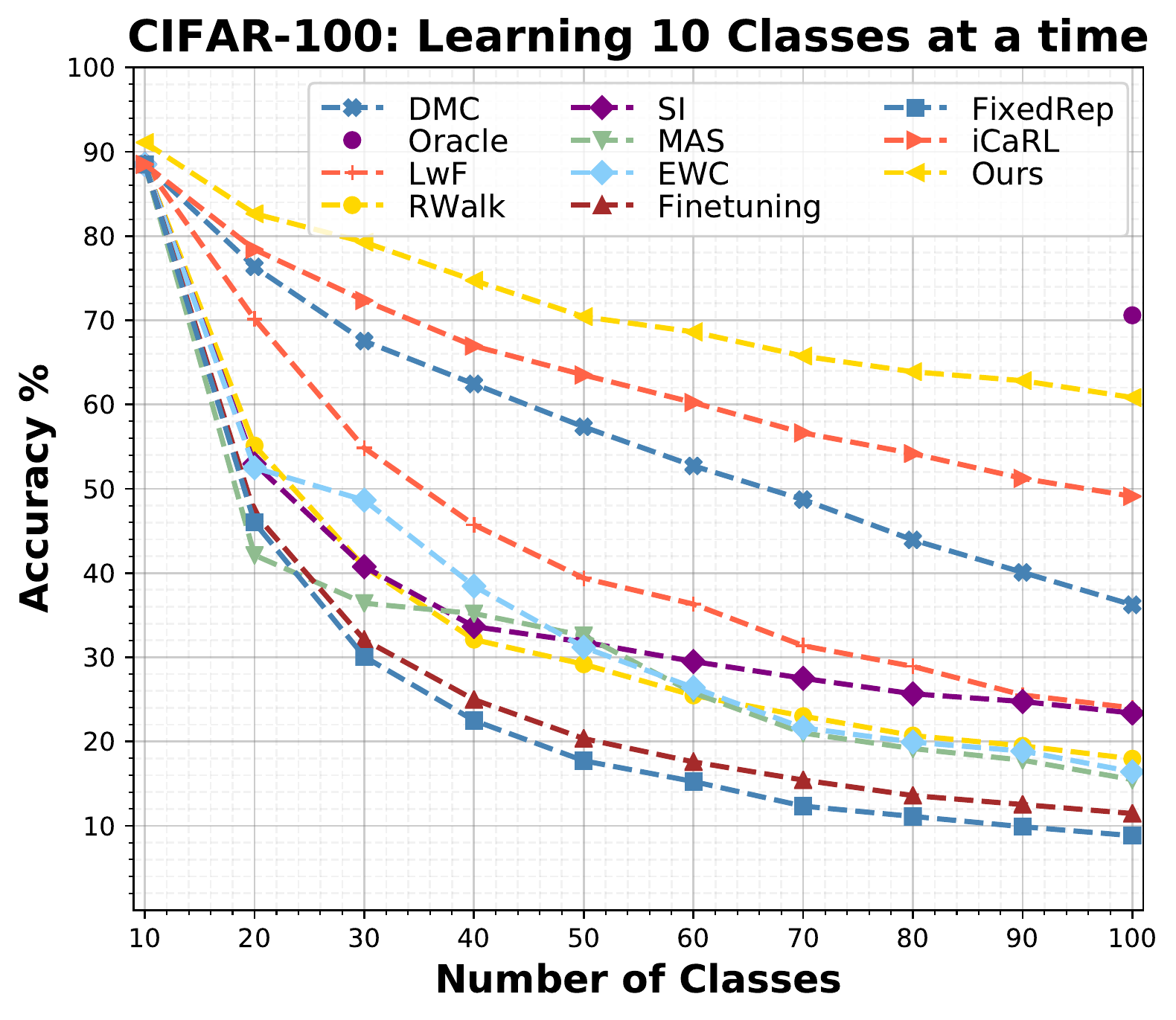}
\includegraphics[width=0.45\textwidth]{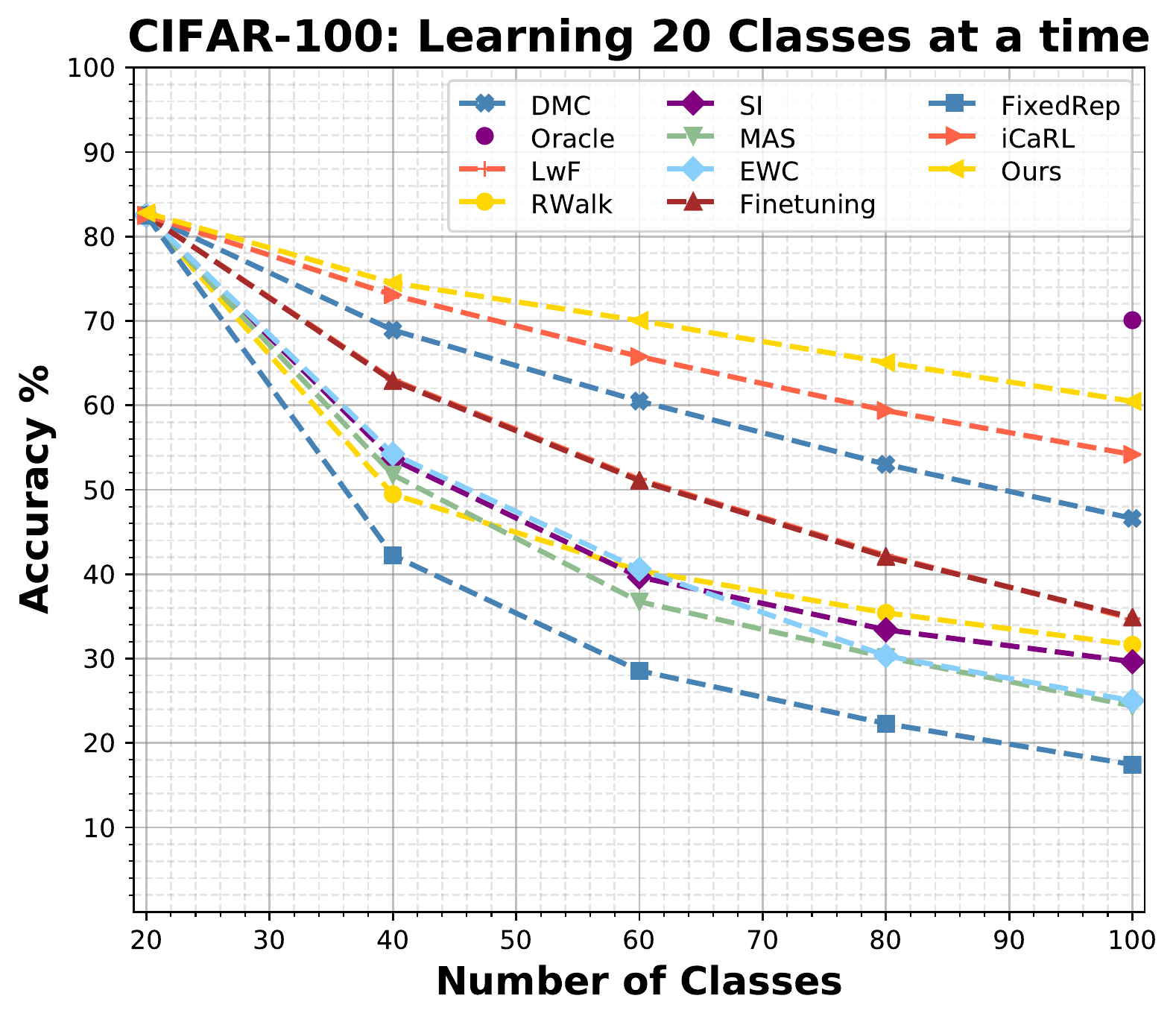}
\includegraphics[width=0.45\textwidth]{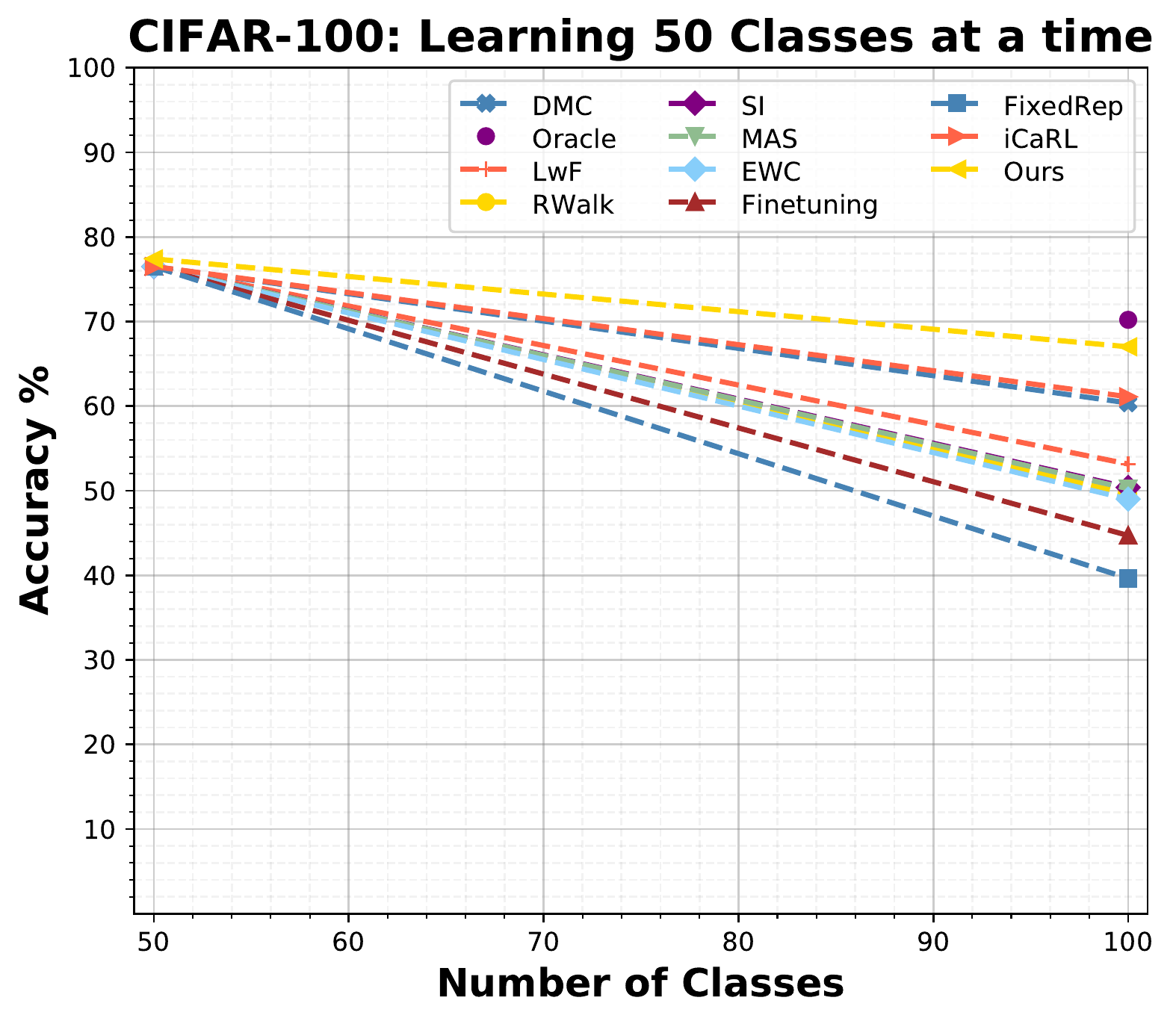}
\caption{ \emph{Results on CIFAR-100}. Evaluations are performed with 20, 10, 5 and 2 tasks (from \emph{left} to \emph{right}). We surpass state-of-the-art results on all four set of experiments. }\label{fig:iCIFAR} 
\end{figure*}

\textbf{Training:}
For the CIFAR100 dataset, we use \texttt{resnet-18} along with max pooling after 5th, 7th blocks and global generalized mean pooling (GeM)~\cite{GeM_pooling} with pooling parameter 3 is used after 9th block. For ImageNet dataset, we use the standard \texttt{resnet-18} architecture as in  \cite{rebuffi2017icarl}. After the GAP layer, a single fully connected layer with weights $\mathbf{W}_{fc} \in \mathbb{R}^{512 \times 100}$  is used as a classifier. For MNIST, a simple 2 layered MLP (with $400$ neurons each), whereas for SVHN \texttt{resnet-18} is used, similar to \cite{Hsu18_EvalCL}.

For each task, we train our model for $50$ epochs using Adam~\cite{kingma2014adam} with $t_e=2$, with learning rate starting from $10^{-3}$ and divided by $2$ after $20,30,40$ epochs. We set the  controller's scaling factor to $\gamma=2.5$ and $\gamma=10$ respectively for CIFAR and ImageNet datasets. We use the ratio between the number of training samples for a task and the fixed number of exemplars as the value for $\gamma$. We fix $M=8, th=0$ through out the experiments. We do not use any weight or network regularization scheme such as dropout in our model. For augmentation, training images are randomly cropped, flipped and rotated ($<10^0$). For each task, we train $N=8$ models in parallel using a Nvidia-DGX-1 machine. These models come from the randomly sampled paths in our approach and may have some parts frozen due to an overlap with previous tasks.

\subsection{Results and Comparisons}

We extensively compare the proposed technique with existing state-of-the-art methods for incremental learning. These include Elastic Weight Consolidation (EWC) \cite{kirkpatrick2017overcoming}, Riemannian Walk (RWalk) \cite{chaudhry2018riemannian}, Learning without Forgetting (LwF) \cite{li2018learning}, Synaptic Intelligence (SI) \cite{zenke2017continual}, Memory Aware Synapses (MAS) \cite{aljundi2018memory}, Deep Model Consolidation DMC \cite{zhang2019class} and Incremental Classifier and Representation Learning (iCARL) \cite{rebuffi2017icarl}. We further evaluate on three baseline approaches: Fixed Representation (\textit{FixedRep}) where the convolution part of the model is frozen and only the classifier is trained for newly added classes, \textit{FineTune} where the complete previously learnt model is tuned for the new data, and \textit{Oracle} where the model is trained on all samples  from previous and current tasks. 

Fig.~\ref{fig:iCIFAR} compares different methods on CIFAR-100 datasets, where we incrementally learn groups of $10$, $20$ and $50$ classes at a time. The results indicate superior performance of the proposed method in all settings. For the case of learning $10$ classes at a time, we improve upon iCARL \cite{rebuffi2017icarl} by an absolute margin of $7.3\%$. Compared with the second best method, our approach achieves a relative gain of $5.3\%$ and $9.7\%$ respectively for the case of incrementally learning $20$ and $50$ classes on CIFAR-100 dataset. For the case of $50$ classes per task, our performance is  only $3.2\%$ below the \textit{Oracle} approach, where all current and previous class samples are used for training. 

\setlength{\tabcolsep}{0.2cm}
\begin{table*}[h]
    \caption{Large Scale experiments on ImageNet-1K and and MS-Celeb-10K show that \ours{} performs favourably against all the state-of-the-art methods. Note that reported task $t$ accuracy is an average of all $1,2, .., t$ tasks.}
    \label{tbl:imagenet}
    \begin{center}
        \begin{tabular}{c l c c c c c c c c c c}
        \toprule[0.4mm]
        \textbf{Datasets} & \textbf{Methods} & \textbf{1} & \textbf{2}& \textbf{3}& \textbf{4}& \textbf{5}& \textbf{6}& \textbf{7}& \textbf{8}& \textbf{9}& \textbf{Final} \\
            \midrule
            \multirow{6}{*}{ImageNet-100/10}
            & Finetuning & 99.3&49.4&32.6&24.7&20.0&16.7&13.9&12.3&11.1&9.9 \\
            & FixedRep & 99.3&88.1&73.7&62.6&55.7&50.2&42.9&41.3&39.2&35.3 \\
            & LwF \cite{li2018learning} & 99.3&95.2&85.9&73.9&63.7&54.8&50.1&44.5&40.7&36.7 \\
            & iCaRL \cite{rebuffi2017icarl} & 99.3&97.2&93.5&91.0&87.5&82.1&77.1&72.8&67.1&63.5 \\
            & Ours & 100.0&97.4&94.3&92.7&89.4&86.6&83.9&82.4&79.4& \ \ \ \ \ \ \ \ \ \ 
            $\textbf{74.1}_{\textcolor{red}{+10.6}}$ \\
            \midrule
            \multirow{5}{*}{ImageNet-1K/10}
            & Finetuning & 90.2&43.1&27.9&18.9&15.6&14.0&11.7&10.0&8.9&8.2 \\
            & FixedRep & 90.1&76.1&66.9&58.8&52.9&48.9&46.1&43.1&41.2&38.5 \\
            & LwF \cite{li2018learning} & 90.2&77.6&63.6&51.6&42.8&35.5&31.5&28.4&26.1&24.2 \\
            & iCaRL \cite{rebuffi2017icarl} & 90.1&82.8&76.1&69.8&63.3&57.2&53.5&49.8&46.7&44.1 \\
            & Ours & 90.2&88.4&82.4&75.9&66.9&62.5&57.2&54.2&51.9& \ \ \ \ \ \ \ \ $\textbf{48.8}_{\textcolor{red}{+4.7}}$ \\
            \midrule
            \multirow{2}{*}{MS-Celeb-10K/10}
            & iCaRL \cite{rebuffi2017icarl} & 94.2&93.7&90.8&86.5&80.8&77.2&74.9&71.1&68.5&65.5 \\
            & Ours & 92.9&94.6&93.6&90.9&87.2&82.3&78.6&77.0&74.0&  \ \ \ \ \ \ \ \ $\textbf{69.2}_{\textcolor{red}{+3.7}}$ \\
            \bottomrule[0.4mm]
        \end{tabular}
    \end{center}
\end{table*}

\begin{table}[htp]
\begin{minipage}{0.48\textwidth}
\vspace{0.5cm}
            
            \captionof{table}{\emph{Comparison on MNIST and SVHN datasets.} Ours is a  memory based approach (denoted by `$*$') that outperforms  state-of-the-art and performs quite close to Joint training (oracle case). 
            }
            \begin{center}
            \scalebox{1.1}{
            \begin{tabular}{l c c}
            \toprule
           \textbf{Methods} & \textbf{MNIST}($A_5$) & \textbf{SVHN}($A_5$)  \\
             \midrule
             Joint training &  97.53\%  &  93.23\% \\ \midrule
              EWC \cite{kirkpatrick2017overcoming} & 19.80\%  &  18.21\% \\
            Online-EWC  \cite{schwarz2018progress}  & 19.77\%  & 18.50\%  \\
            SI   \cite{zenke2017continual}     & 19.67\%  &  17.33\% \\
            MAS  \cite{aljundi2018memory}     & 19.52\%  &  17.32\% \\
            LwF  \cite{li2018learning}     & 24.17\%  &  - \\
            \midrule
            GEM$^*$ \cite{lopez2017gradient}       &   92.20\%  &  75.61\% \\
             DGR$^*$ \cite{shin2017continual}       &   91.24\%  & -  \\
            RtF$^*$ \cite{van2018generative}       &   92.56\%  & - \\
            \midrule
             {Ours}$^{*}$  &   \textbf{96.16}\%  & \textbf{90.83}\%  \\
            \bottomrule
            \end{tabular}}
            \end{center}
            \label{tbl:mnsit_svhn}
\end{minipage}  
\end{table}

Fig.~\ref{tbl:imagenet} compares different methods on ImageNet dataset. The results show that for experimental settings consistent with iCARL \cite{rebuffi2017icarl}, our proposed method achieves a significant absolute performance gain of $10.6\%$ compared with the existing state-of-the-art~\cite{rebuffi2017icarl}. Our experimental results indicate that commonly used technique of fine-tuning a model on new classes is clearly an inferior approach, and results in catastrophic forgetting.
Table~\ref{tbl:mnsit_svhn} compares different methods on MNIST and SVHN datasets following experimental setting of \cite{Hsu18_EvalCL}. The results show that \ours{}, surpasses all previous methods with a margin of $4.3\%$ and $13.3\%$ respectively for MNIST and SVHN datasets. The results further indicate that the methods which do not use a memory perform relatively lower.




\subsection{Ablation Studies and Analysis}\label{sec:ablation}
In this section, we report ablation experiments and analyze the behaviour of our approach under different configurations of hyper-parameters in our approach. 

\begin{figure*}
    \centering
    \begin{subfigure}{.245\textwidth}
        \centering
        \includegraphics[scale=0.28]{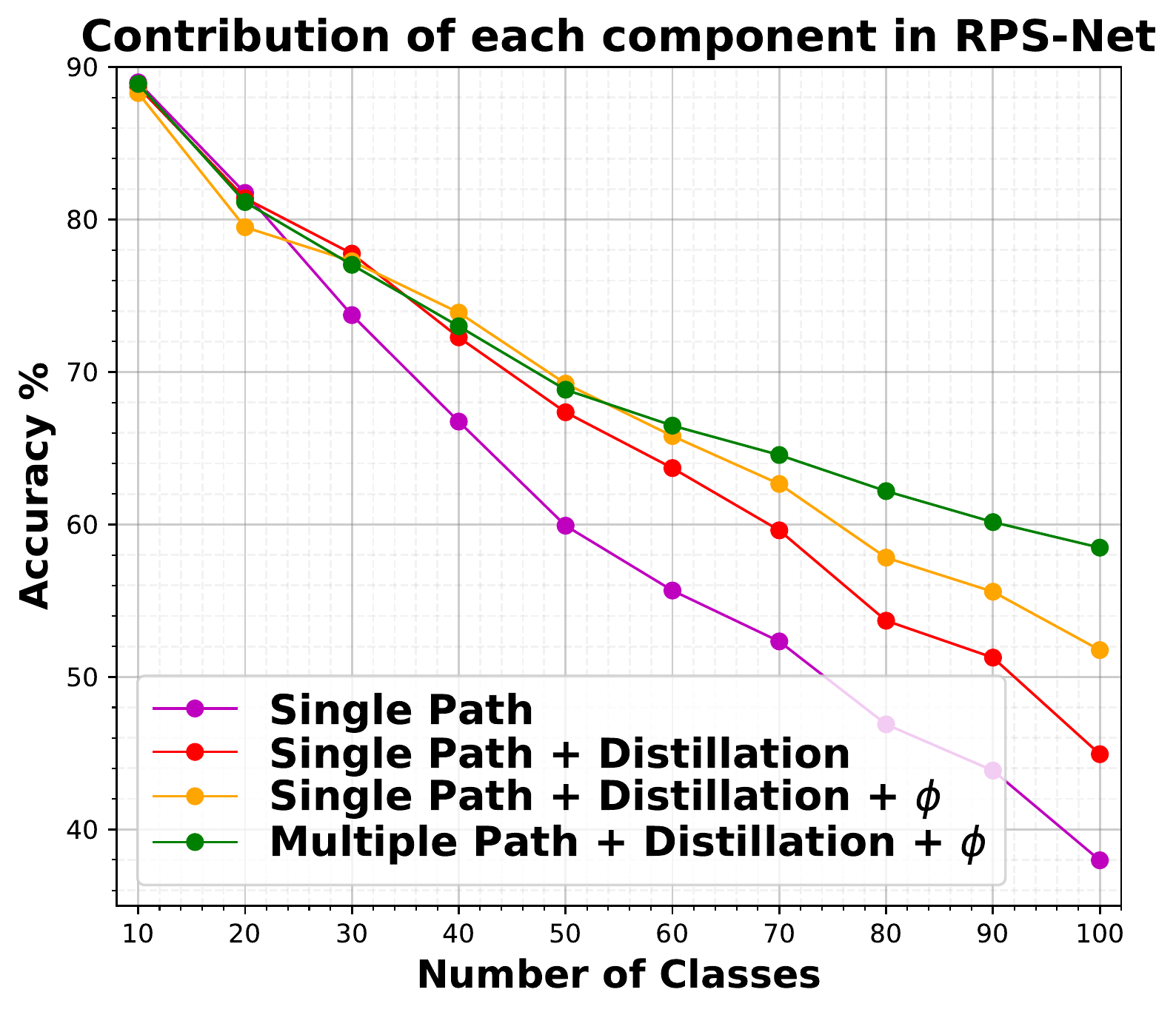}
        \caption{Individual Contributions}\label{fig:indv_contrib}
    \end{subfigure}
    \begin{subfigure}{.245\textwidth}
        \centering
        \includegraphics[scale=0.28]{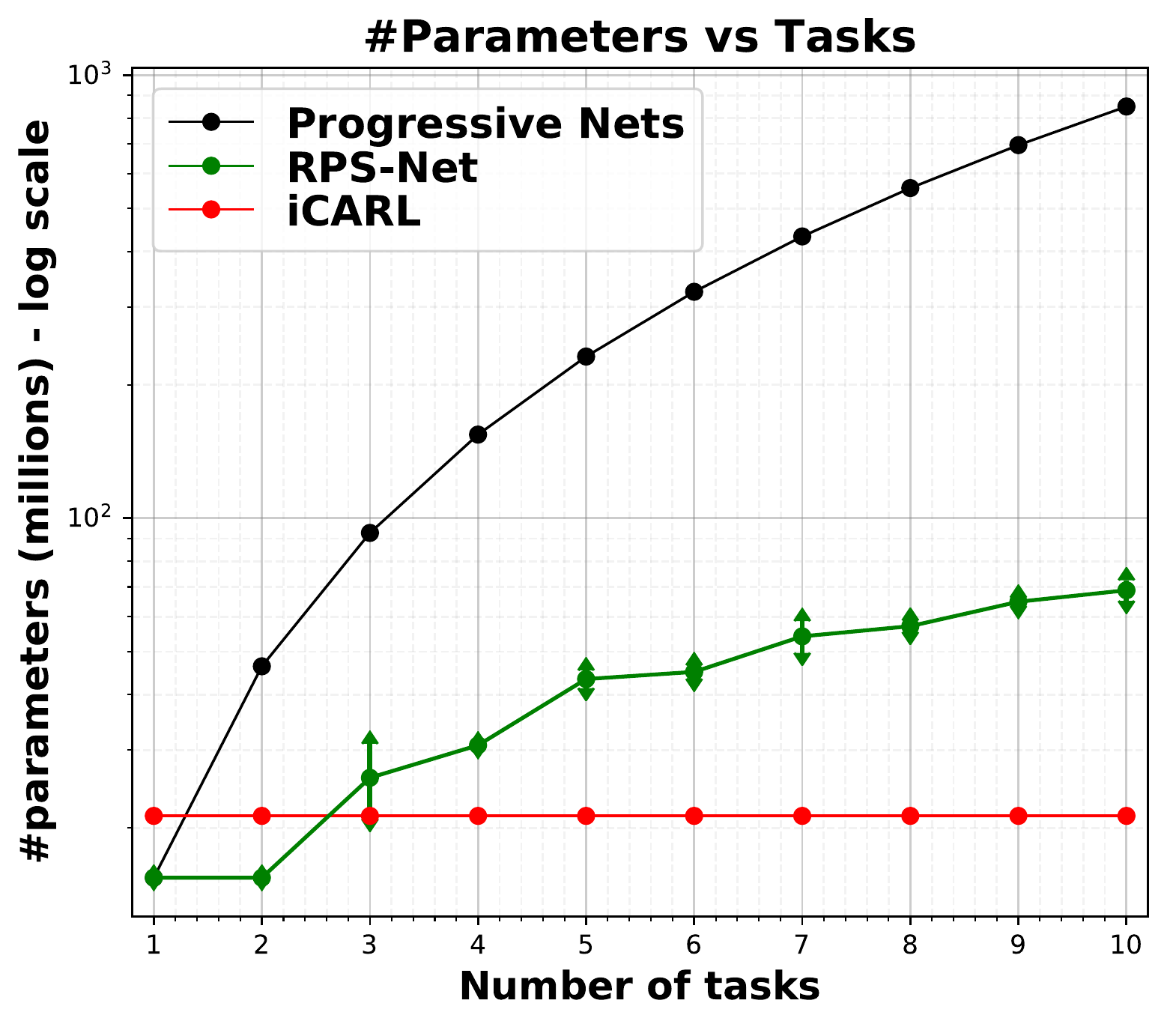}
        \caption{$\#$Parameters}\label{fig:params}
    \end{subfigure}
    \begin{subfigure}{.245\textwidth}
        \includegraphics[scale=0.28]{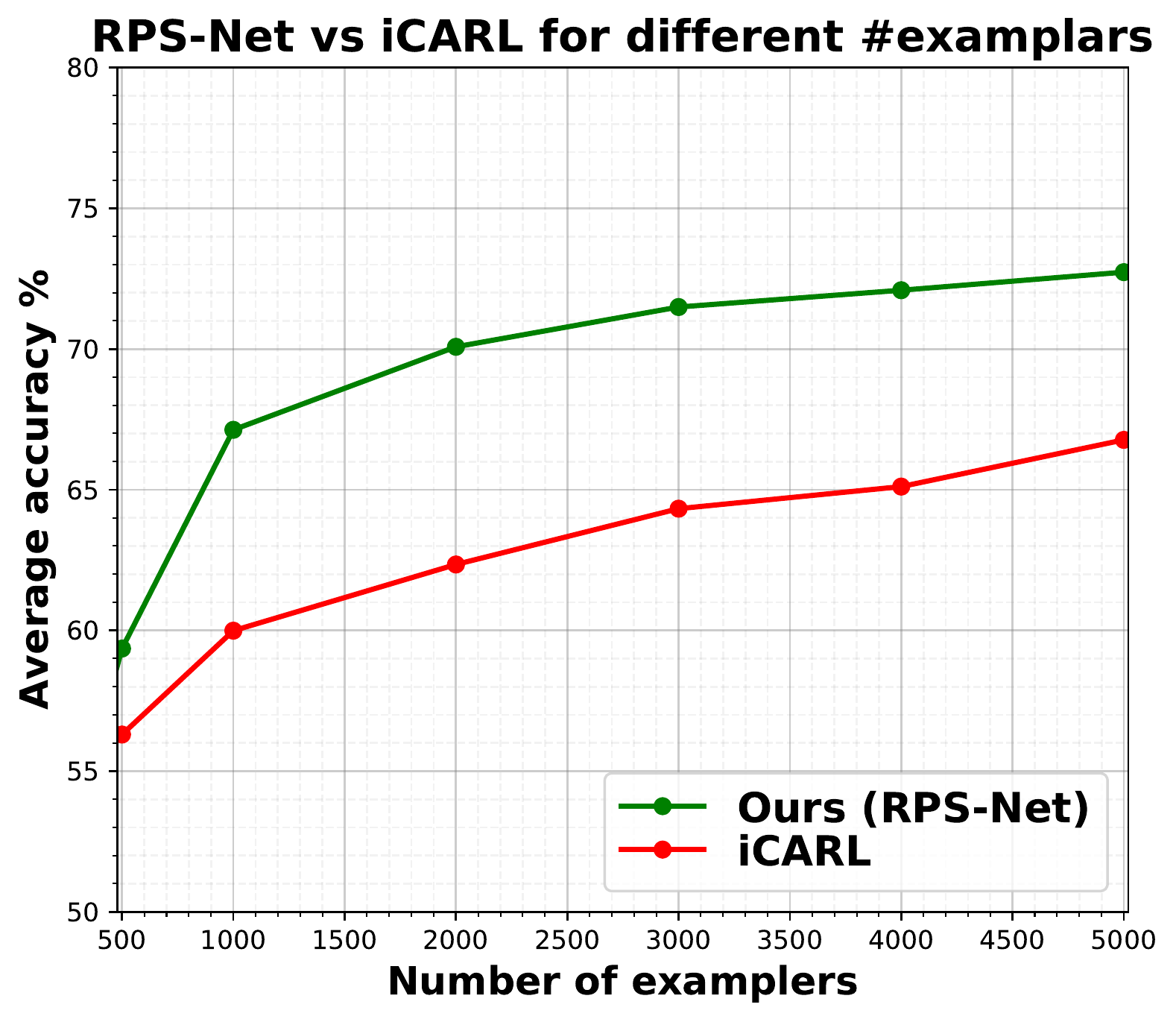}
        \caption{$\#$Exemplars}\label{fig:exemplars}
    \end{subfigure}
    \begin{subfigure}{.245\textwidth}
        \includegraphics[scale=0.28]{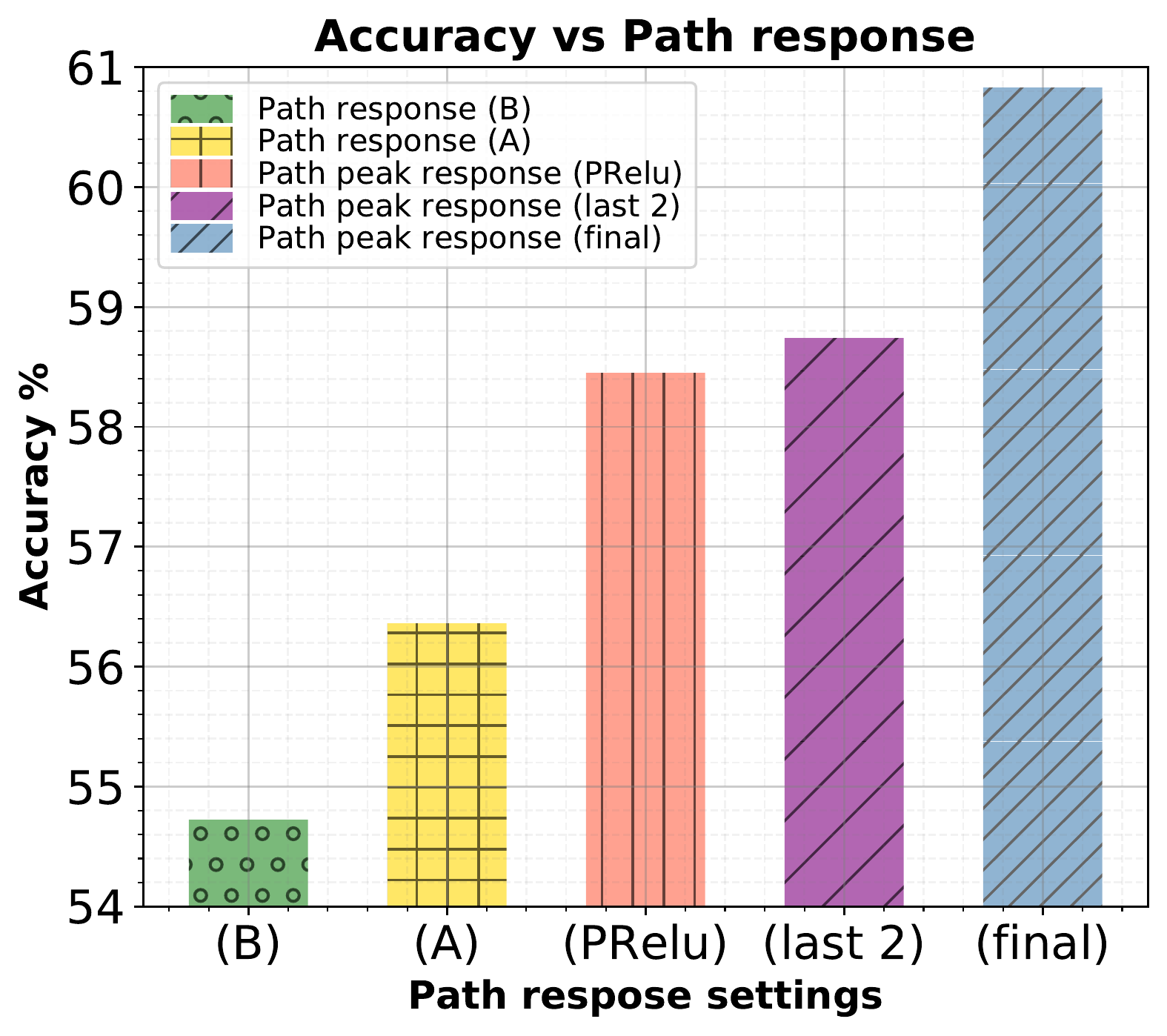}
        \caption{Attention functions}\label{fig:gamma_mapping} 
    \end{subfigure}
    \caption{\emph{Detailed experimental analysis of \ours{}.} (a) Ablation experiments to study the contribution from individual components of our approach, (b) The effect of number of tasks on the overall parametric complexity of \ours{} versus iCARL \cite{rebuffi2017icarl} and Progressive Networks \cite{rusu2016progressive}. (c) The effect of changing the number of exemplars on ours and iCARL \cite{rebuffi2017icarl}. (d) The comparison between different attention functions used to re-weight the path responses in \ours{}. The classification accuracy on CIFAR100 with 10 tasks is reported. }
    \label{fig:analysis}
\end{figure*}

\textbf{What really matters?} Fig.~\ref{fig:indv_contrib} studies the impact of progressively integrating individual components of our \ours{}. We begin with a simple baseline model with a single path that achieves $37.97\%$ classification accuracy on CIFAR100 dataset. When distillation loss is used alongside the baseline model, the performance increases to $44.93\%$. The addition of our proposed controller  $\phi (k, \gamma)$ in the loss function further gives a significant boost of $+6.83\%$, resulting in an overall accuracy of $51.76\%$. Finally, the proposed multi-path selection algorithm along with above mentioned components increases the classification accuracy up to $58.48\%$. This demonstrates that our two contributions, controller and multi-path selection, provide a combined gain of $13.6\%$ over baseline + distillation. Note that these experimental results are obtained without our proposed dynamic path switching rule and the attention function whose effect is extensively explore later in this section.

\textbf{Model Size Comparison:} Although our approach dynamically increases the network's capacity to allow learning new tasks, it is important to note that the parametric complexity remains bounded for a large number of tasks. Fig.~\ref{fig:params} compares total parameters across tasks for Progressive Nets \cite{rusu2016progressive}, iCARL \cite{rebuffi2017icarl} and our \ours{} on CIFAR100. Our model effectively reuses previous parameters, and the model size does not increase significantly with tasks. After 10 tasks, {RPS-Net} has 72.26M parameters on average, compared with iCARL  (21.3M) and Progressive Nets (932.84M). This shows that in RPS-Net, learnable parameters increase logarithmically, while for Progressive Nets they increase quadratically.

\textbf{Varying Number of Exemplars:} Ours is a memory based approach that keeps a small set of exemplars for memory replay at later stages to avoid catastrophic forgetting. Generally, a larger set of exemplars should help improve the performance. It is therefore important to study how the RPS-Net performs for different sizes of exemplar set. Fig.~\ref{fig:exemplars} compares RPS-Net with the best existing method (iCARL) for various memory budgets of exemplars on CIFAR100 dataset. Overall, an improvement is observed as the memory budget is increased for both approaches, but our proposed RPS-Net consistently performs better across all budgets. 

\textbf{Attention function:} In this section, we study several other attention mechanisms in comparison to our proposed peak response function (see~Fig.~\ref{fig:gamma_mapping}). To this end, we try path mean response function, in which we simply take the average of the activations after a convolution module in the last layer which results in a scalar value and can be directly used to re-weight the path responses. We also use an attention function with one convolution layer and a shared fully connected layer ($\texttt{path response}-(A)$). First a convolution layer  converts $w \times w \times c$ tensor to $w \times w \times 1$ matrix, and $M$ such feature maps are concatenated to form a $w  \times w \times M$ tensor and fed into a shared fully connected layer to output a $M$ dimensional vector and its corresponding dimensions are used to re-weight the $M$ output tensors. In the cases where a path is not activated, a zero tensor is used. Similarly, for $\texttt{path response}-(B)$, maximum activation values of the tensor along the channels are used to get $c$ features per path (in this case, last layer $c=512$), and a shared fully connected layer is used to get $M$ re-weight coefficients.

\begin{figure*}[!htb]
    \centering
    \begin{subfigure}{.245\textwidth}
        \centering
            \includegraphics[width=0.99\textwidth]{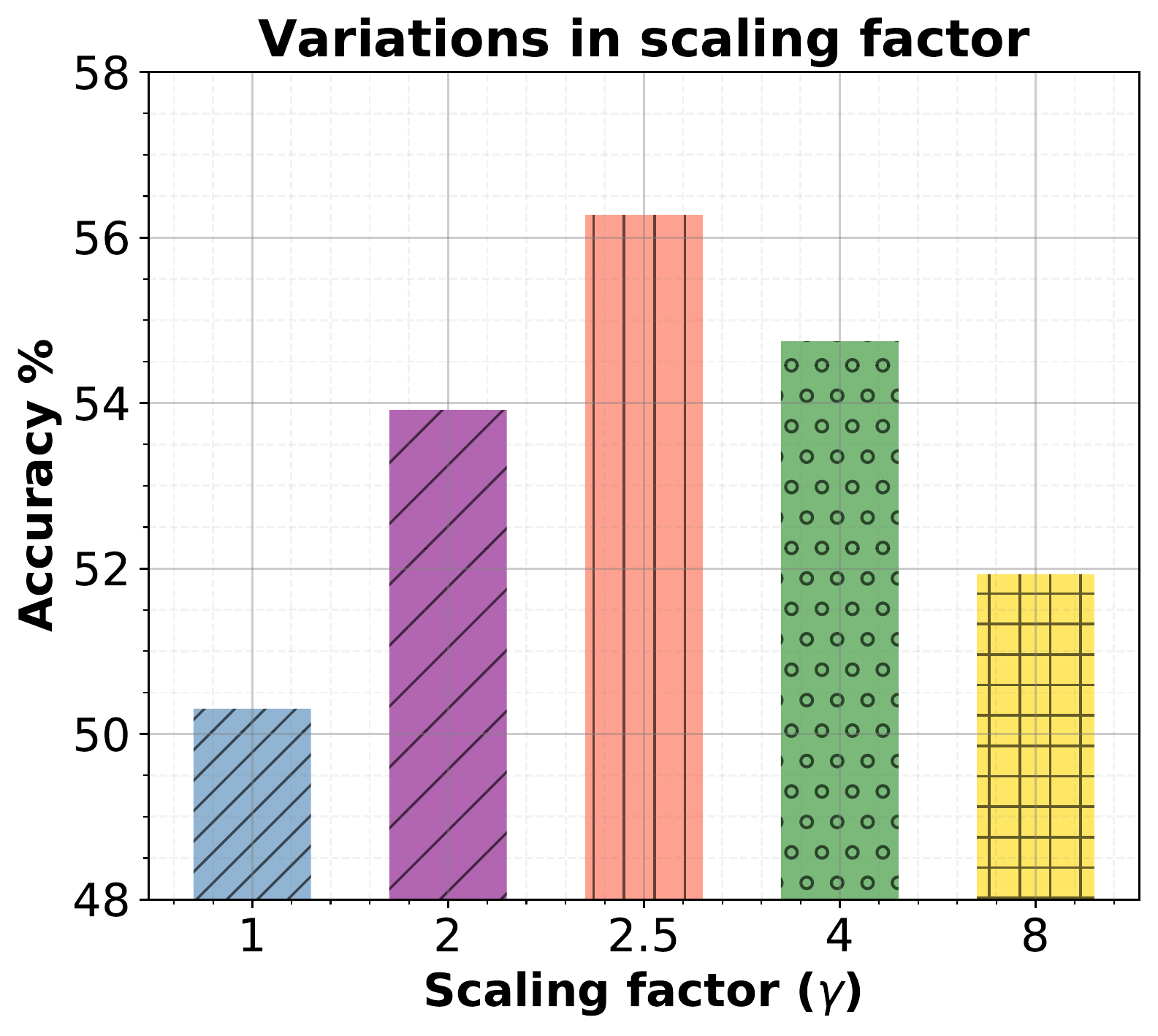}
        \caption{Parameter $\gamma$}\label{fig:gamma}
    \end{subfigure}
    \begin{subfigure}{.245\textwidth}
        \centering
        \includegraphics[width=0.99\textwidth]{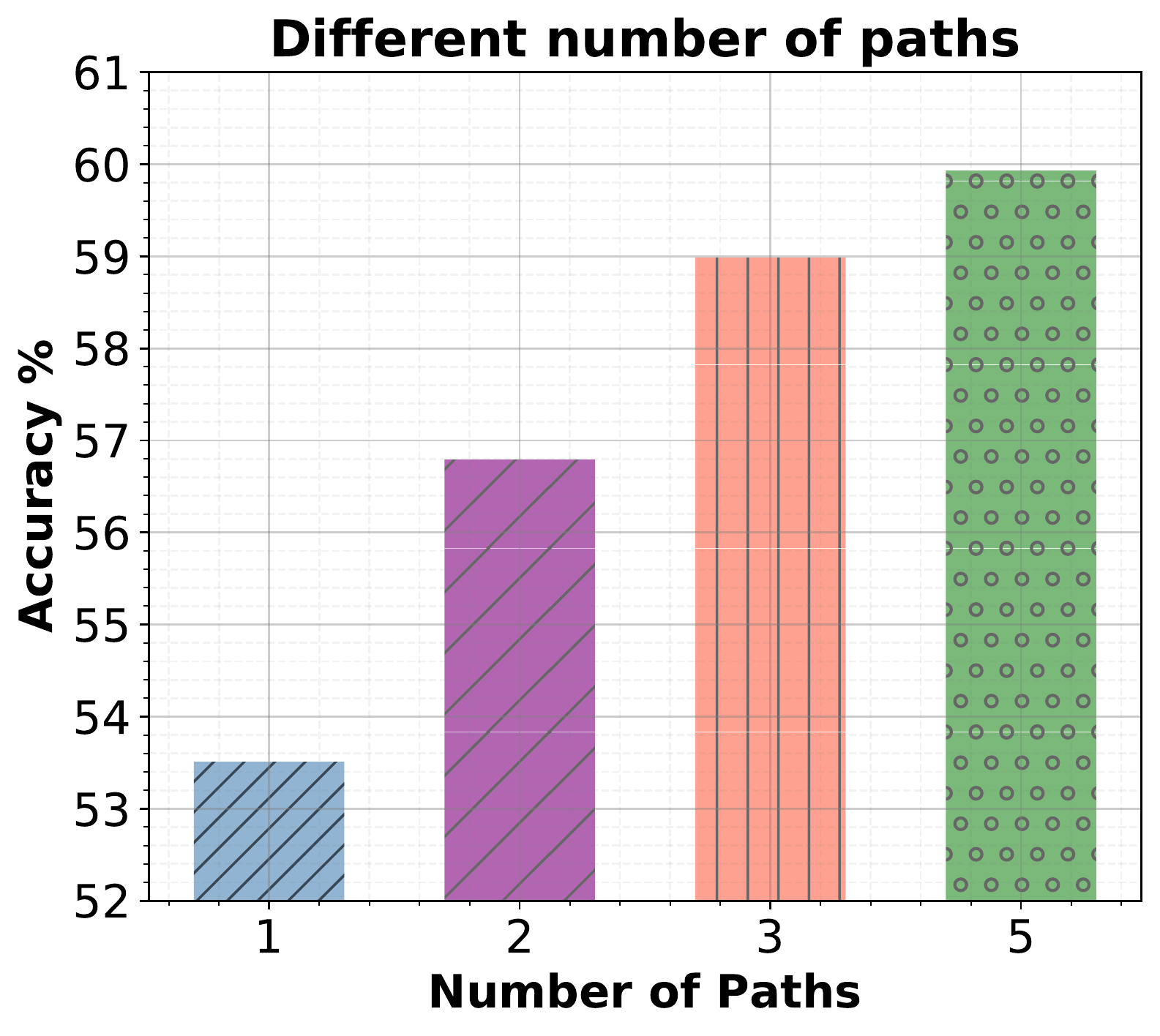}
        \caption{Parameter $J$}\label{fig:j_var}
    \end{subfigure}
    \begin{subfigure}{.245\textwidth}
        \centering
        \includegraphics[width=0.99\textwidth]{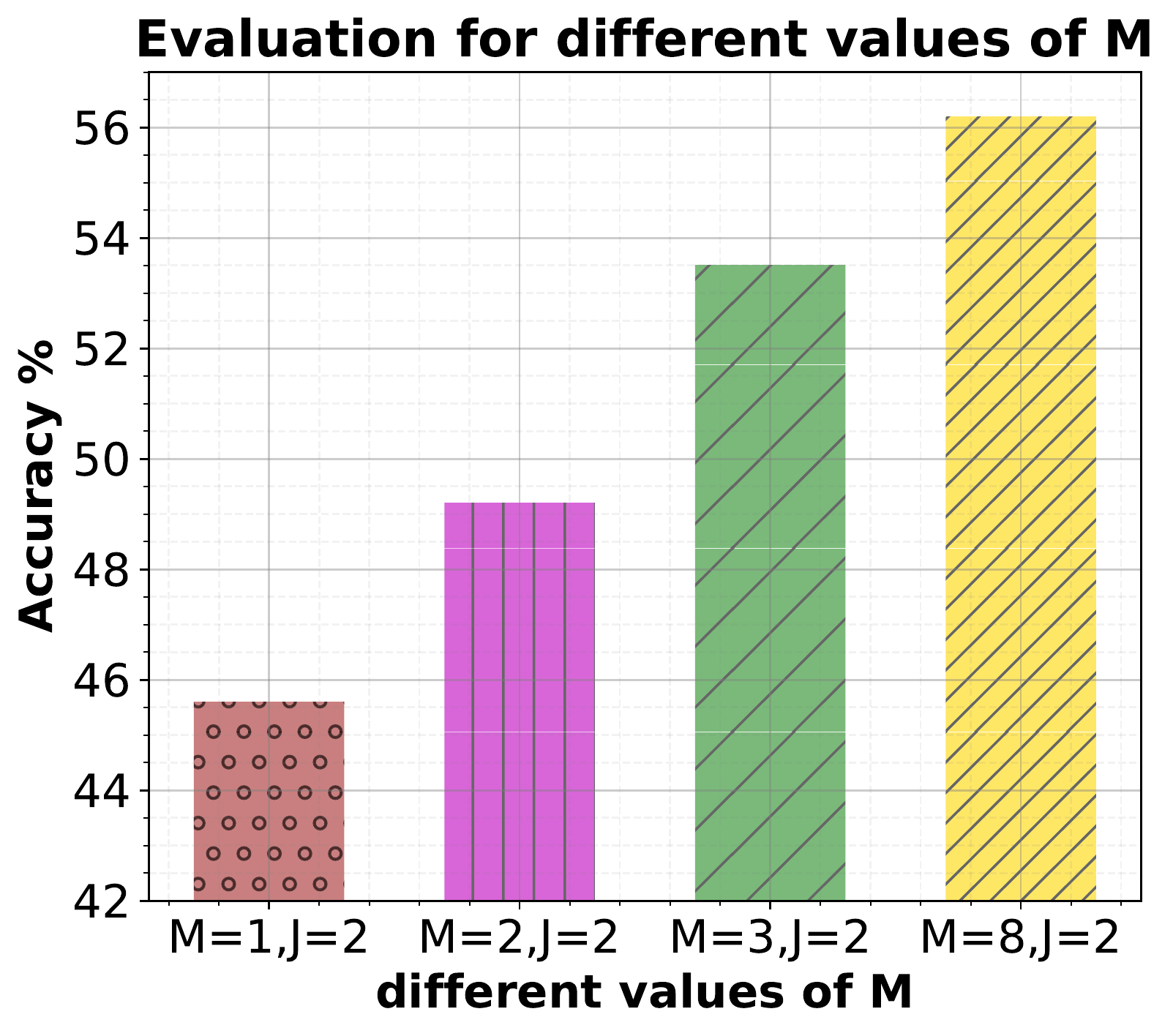}
        \caption{Parameter $M$}\label{fig:m_var}
    \end{subfigure}
    \begin{subfigure}{.245\textwidth}
        \centering
        \includegraphics[width=0.86\textwidth, angle=-90]{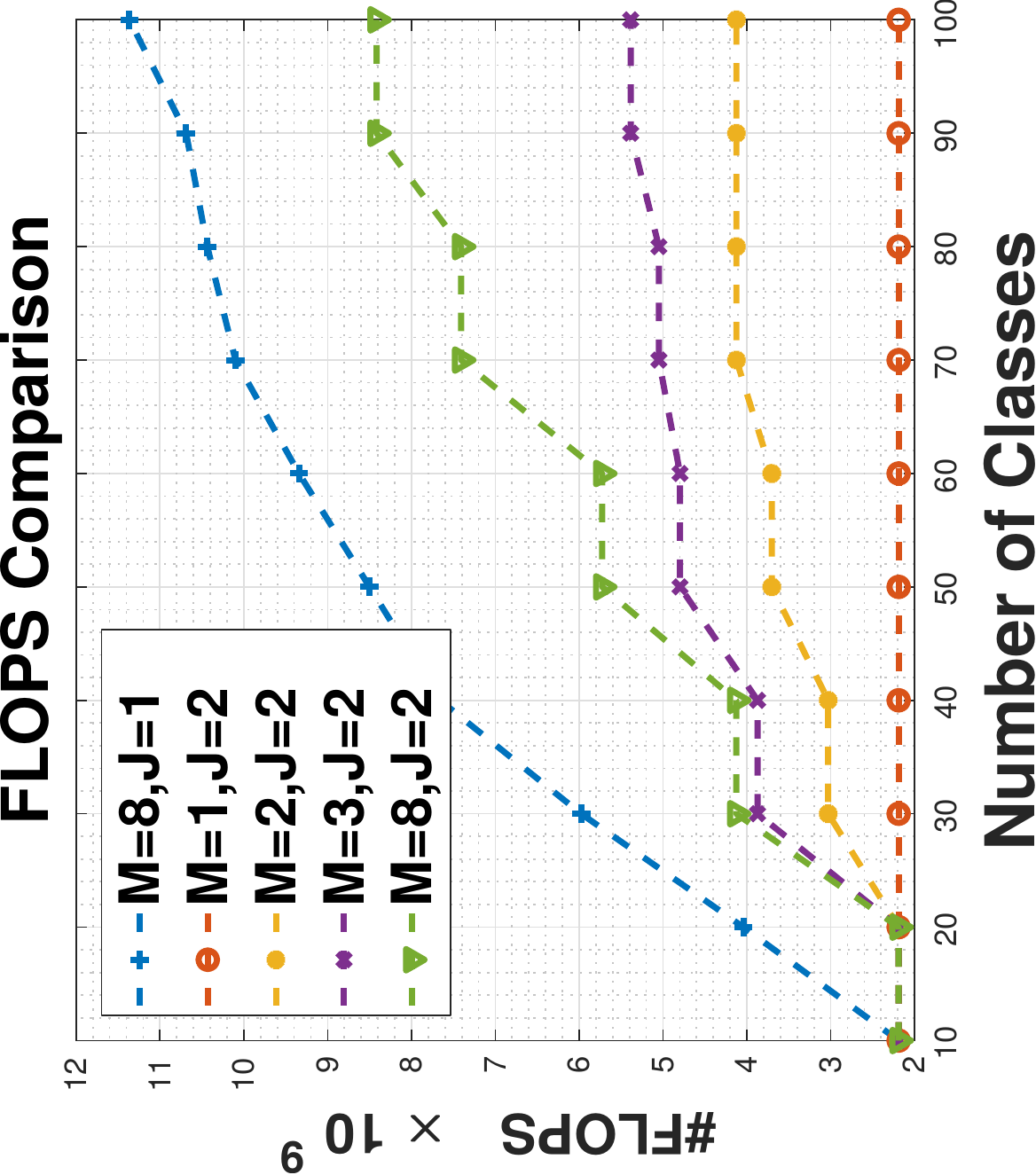}
        \caption{$M,J$ with FLOPS}\label{fig:flops}
    \end{subfigure}
   
    \caption{\emph{Parameter Sensitivity Analysis.}  Performance trend with different settings of parameters $\gamma$, $J$ \& $M$ are reported (from \textit{left to right}). The effect of changing parameters $M$ and $J$ on the computational complexity (in terms of FLOPS) of \ours{} is shown in the right-most plot. }
    \label{fig:ablation}
\end{figure*}

\textbf{Scaling Factor $\gamma$:} It controls the equilibrium between cross-entropy and distillation losses (or the balance between new and old tasks). In Fig.~\ref{fig:gamma}, for smaller $\gamma$, the network tends to forget old information while learning the new tasks well and vice versa. For example, when $\gamma=1$ (same as loss function used in iCaRL~\cite{rebuffi2017icarl}) the performance drops, showing the model is not at its equilibrium state. On the other hand, $\gamma = 8$ also shows a drop in performance towards the later tasks ($51\%$ at task $10$). Empirically, we found the optimal value for $\gamma=2.5$, to keep the equilibrium till last tasks. In this ablation experiment, we keep the network configuration same for all cases, thus we manually change the paths every two task. This is to remove the effect of dynamic path switching which change the effect of $\gamma$. For example for a small value of $\gamma$, network may not remember all the past information, thus a single path can be used many times, while for a higher value of $\gamma$, paths will saturate faster.

\textbf{Varying Blocks and Paths:} 
 One of the important restriction in \ours{} design is the networks' capacity, upper-bounded by $M {\times} L$ modules. As proposed in the learning strategy, a module is trained only once for a path. Hence, it is interesting to study whether the network saturates for a high  number of tasks. To analyze this effect, we change the parameters $M$ and $J$. Here, M is the number of modules in a layer, and $J$ is the number of tasks a path is trained without switching. Here we do not use our dynamic path selection strategy since we cannot control when switching happens in that case.
 Our results with varying $M$ are reported in Fig.~\ref{fig:m_var}, which demonstrate that the network can perform well even when all paths are saturated. This effect is a consequence of our residual design where the last classification layer and skip connections are always trained, thus helping to continually learn new tasks even if the network is saturated. If saturation occurs, the model has already seen many tasks, therefore it learns generalizable features that can work well for future tasks with the help of residual signal (carrying complementary information) via skip connections and adaptation of the final classification layer weights.
 
 \begin{figure*}[t]
\center
\includegraphics[width=0.37\textwidth]{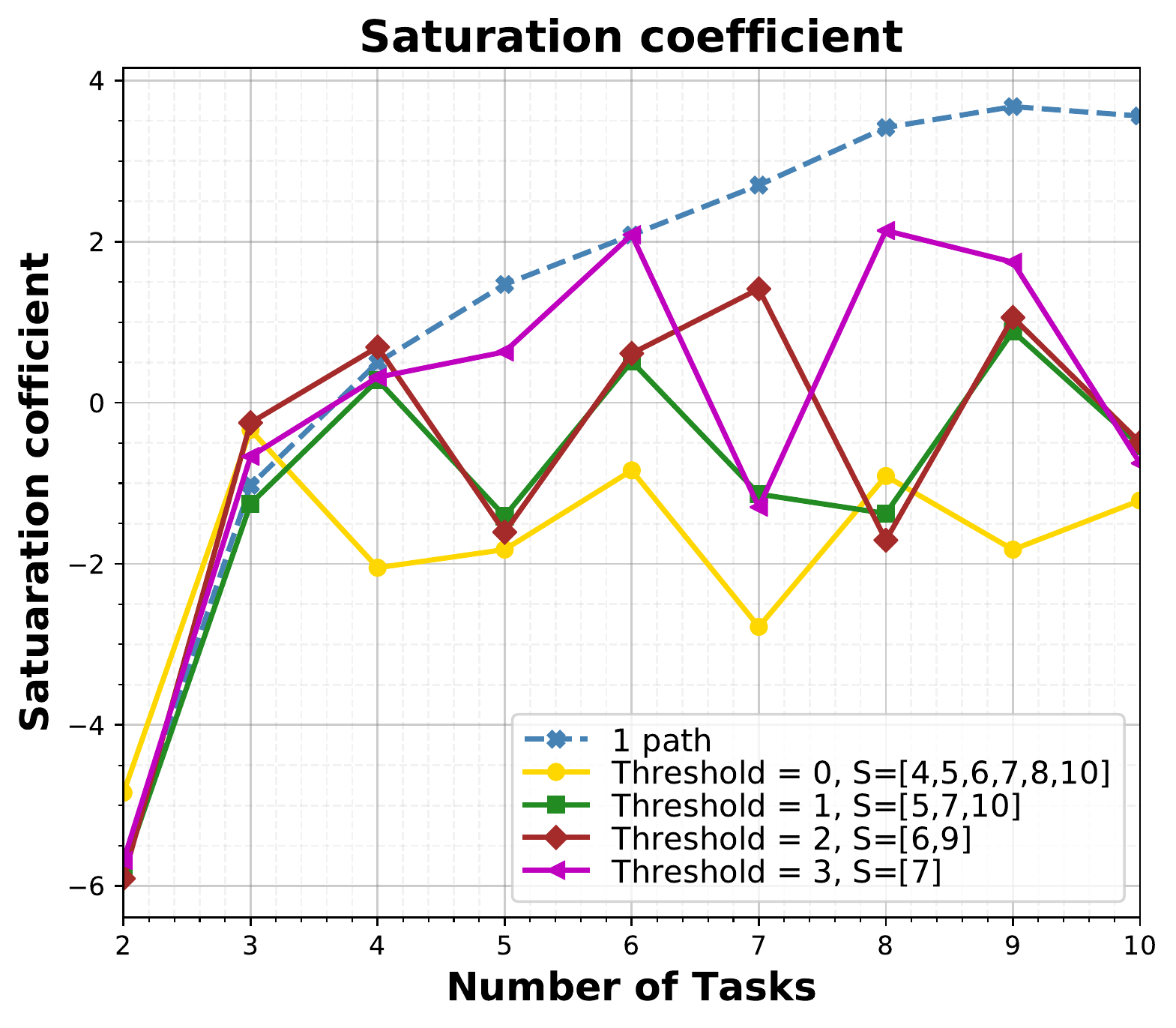}\qquad
\includegraphics[width=0.37\textwidth]{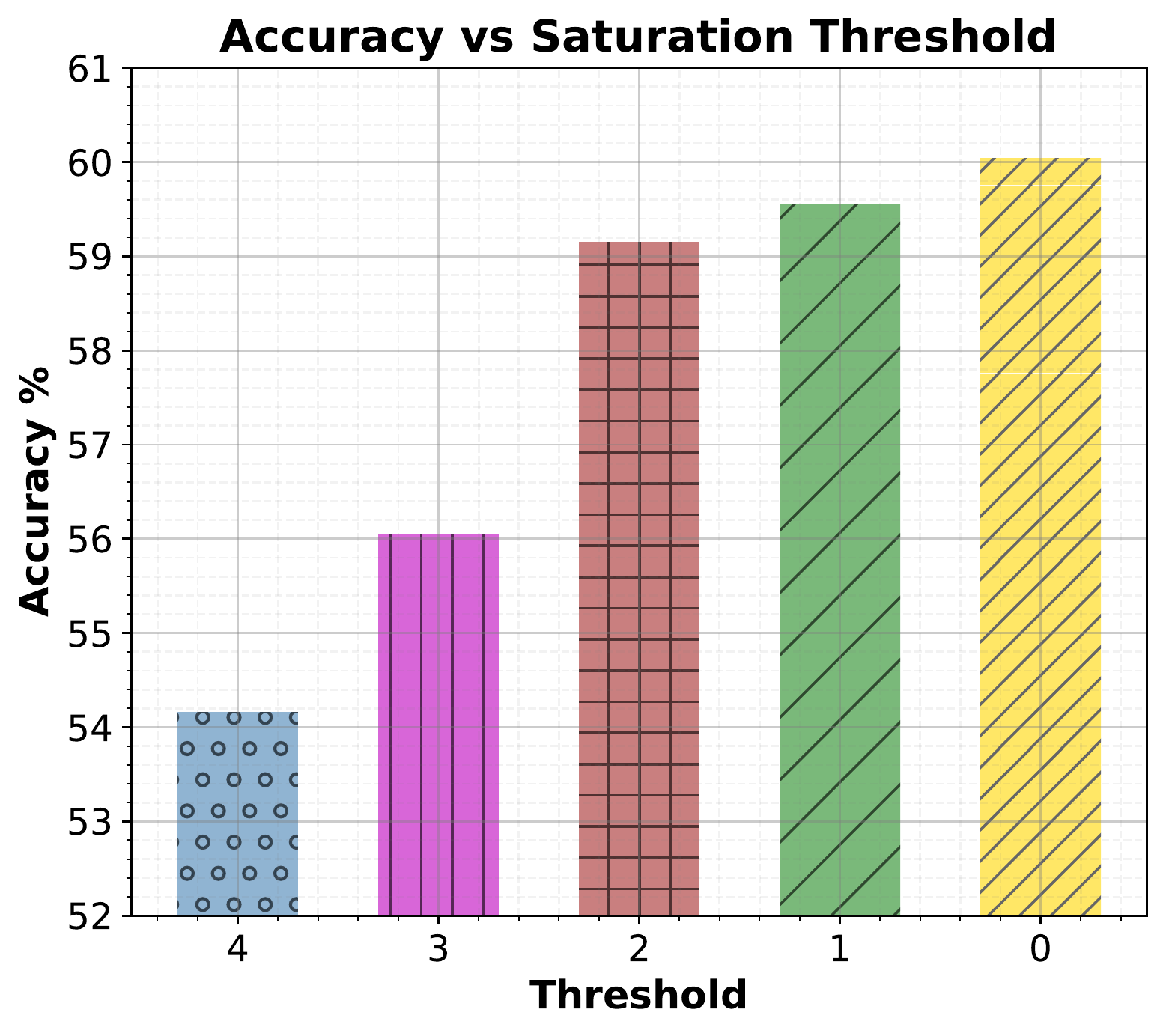}
\caption{\emph{Trend for saturation coefficient and model performance under different path switching rules.}  \emph{Left}, we can see a sawtooth type pattern, this is mainly because once a saturation coefficient passes the threshold value, the network will chose a new path (with possibly several untrained modules). Thus the saturation coefficient will drop with a new path. Dashed line show the case when paths are changed manually, while solid lines denote the case when paths are changed automatically with a given threshold on saturation coefficient. \emph{Right}, the plot shows that higher number of paths  results in better performance. However, too many paths will introduce more computational overhead and memory requirements. Using an automatic path switching rule (based on a threshold for saturation coefficient), our method finds the best place to switch to a new task, and therefore balance the trade-off between performance and computational complexity. }\label{fig:saturation} 
\end{figure*}
 
In Fig.~\ref{fig:j_var}, we illustrate results with varying paths (paths $\propto \frac{1}{J}$) in the network. We note that learning a high number of paths degrades performance as the previously learned parameters are less likely to be effectively reused. On the other hand, we obtain comparable performance with fewer paths (e.g., 2 for CIFAR-100).

\textbf{FLOPS comparison:} As the number of tasks increase, the network's complexity grows. As shown in Fig.~\ref{fig:flops}, with different configurations of modules and paths, the computational complexity of our approach scales logarithmically. This proves that the complexity of \ours{} is bounded by  $\mathcal{O}(log(\#task))$. This is due to the fact that the overlapping modules increase as the training progresses. Further, in our setting we chose new paths after every $J>1$ tasks. Hence, in practice our computational complexity is well below the worst-case logarithmic curve. For example with a setting of $M{=}2,J{=}2$ the computational requirements reduces by $63.7\%$ while achieving the best performance. We also show that even when a single path is used for all the tasks ($M{=}1$), our model achieves almost the same performance as state-of-the-art with constant computational complexity.

\textbf{Saturation coefficient:}
One of the important hyper-parameter of \ours{} is $\mu_{sat}$, which controls the trade-off between the network resource allocation and performance. Compared to the case when a fixed rule is used to switch paths (e.g., after every 2 tasks resulting in 5 paths), an adaptive rule based on network saturation with $th=2$ helps us achieve similar performance with only 3 paths (see~Fig.~\ref{fig:saturation}). This illustrates that our proposed dynamic path switching scheme that intelligently expands model capacity can help in significant reduction of the computational requirements in \ours{}. Specifically, the best performance is achieved by changing paths for every two tasks, manually. Further, we can surpass iCaRL~\cite{rebuffi2017icarl} performance with only two different paths ($th=3$). This shows that, not only we able to learn the places where we need to jump the paths, this helps in maximum utilization of available resources. 

\begin{figure}[t]
\begin{center}
         \includegraphics[width=0.4\textwidth]{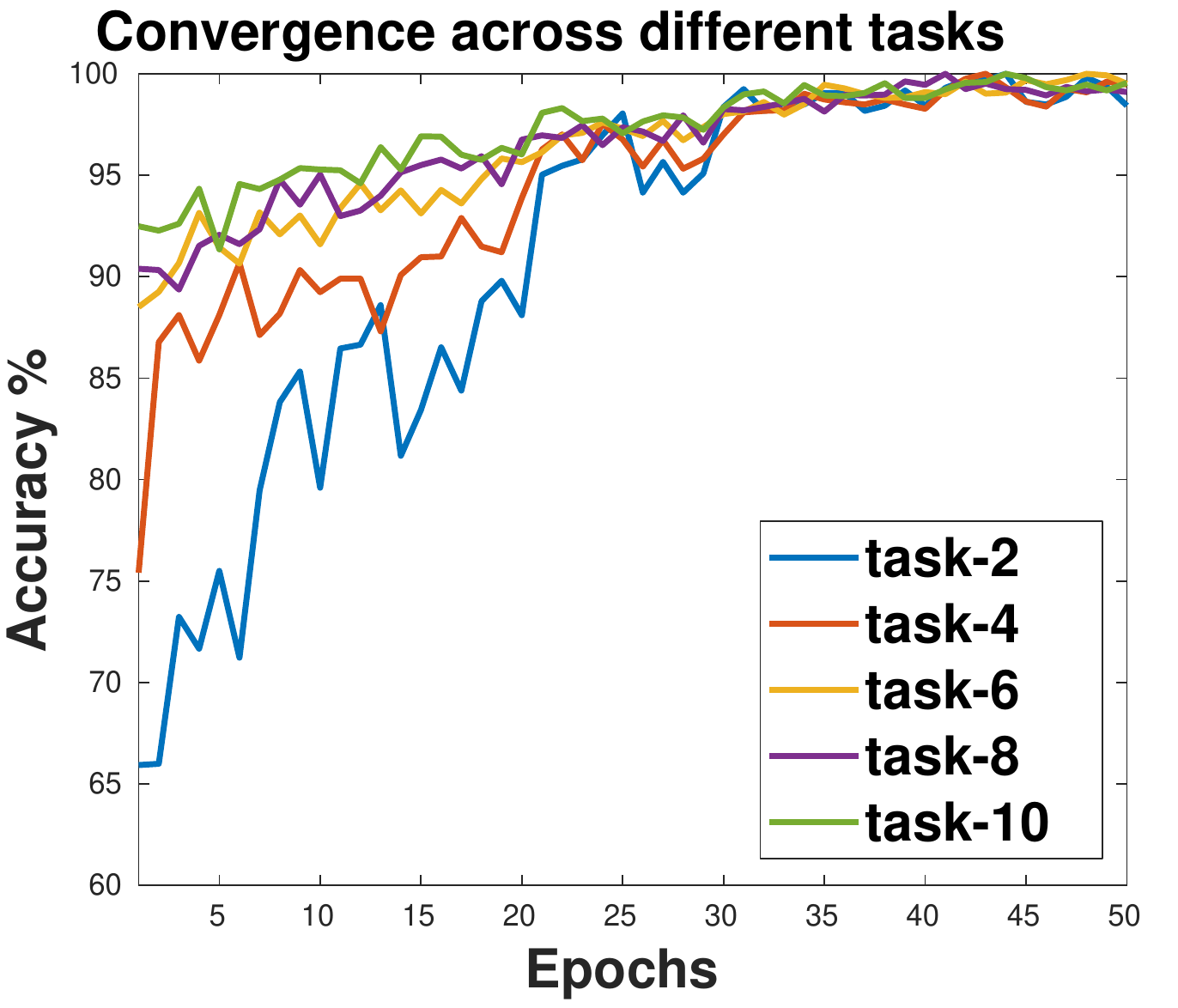}
\end{center}
\caption{\emph{Forward Transfer}. \ours{} converges fast for the final tasks, showing forward transfer as the learning progresses.}
\label{fig:conv}
\end{figure}

\begin{figure*}[t]
\center
\includegraphics[trim={0cm 0cm 0.94cm 0cm},clip,scale=0.6, angle=-90]{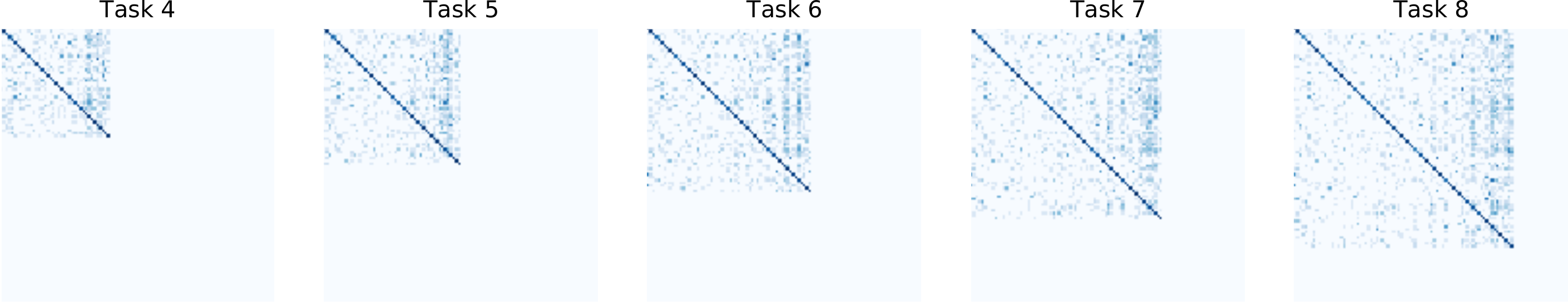}
\caption{\label{fig:cfmatrix} \emph{Bakward Transfer.} Confusion matrices  
over 10 incremental tasks on CIFAR-100, showing backward knowledge transfer. For better visualization, a transformation of log(x+1) is applied to the confusion matrices. 
}
\end{figure*}

\textbf{Forward Transfer:}
The convergence trends shown in Fig.~\ref{fig:conv} demonstrate the forward knowledge transfer for \ours{}. We can see that for task-$2$, the model takes relatively longer to converge compared with task-10. Precisely, for the final task, the model achieves $95\%$ of the total performance within only one epoch, while for the second task it starts with $65\%$ and takes up-to $20$ epochs to achieve $95\%$ of the final accuracy. This trends shows the faster convergence of our model for newer tasks
This effect is due to residual learning as well as overlapping module sharing in \ours{} design, demonstrating its forward transfer capability.

\textbf{Backward Transfer:}
Fig.~\ref{fig:cfmatrix} shows evolution of our model with new tasks. We can see that the performance of the current task (\textit{k}) is lower than the previous tasks (\textit{<k}). Yet, as the model evolves, the performance of task $k$ gradually increases. This demonstrates models' capability of backward knowledge transfer, which is also reflected in biological aspects of human brain. Specifically, hippocampus in human brain accomplishes fast learning which is later slowly consolidated with the slow learning at neocortex~\cite{DBLP:journals/corr/abs-1802-07569}. In Fig.~\ref{fig:cfmatrix}, we can see the pattern of slow learning, with the performance on new tasks gradually maturing.

\textbf{Comparison with a Genetic Algorithm:} We compare our random selection approach with a genetic algorithm i.e., Binary Tournament Selection (BTS) \cite{miller1995genetic} for 25 maximum generations, on MNIST with 5 tasks (each of 2 classes), using a simple 2 layer (100 neurons) MLP with $M=8, J=1$. On 5 runs, our proposed random selection achieves an average accuracy of $96.52\%$ vs BTS gets $96.32\%$. For same time complexity as ours, BTS has an average accuracy of $71.24\%$ for the \textit{first} generation models. For BTS to gain similar performance as our random selection, it needs an average of $10.2$ generations (much higher than the number of random paths), hence BTS has more compute complexity. Although BTS is a simple genetic algorithm and more sophisticated genetic algorithms may outperform random selection, but likely with a high compute cost, which is not suitable for an incremental classifier learning setting having multiple tasks.

\begin{table}[h]
\begin{minipage}{1\columnwidth}
\vspace{0.5cm}
            \captionof{table}{{\emph{Comparison between RPS-Net and \ours{}.} Ours is a modified version of RPS-Net with adaptive path selection strategy.}}
            \begin{center}
            \begin{tabular}{l c c c}
            \toprule
           \textbf{Dataset} & \textbf{Tasks} & \textbf{RPS-Net} & \textbf{Adaptive RPS-Net}  \\
             \midrule
            CIFAR100 & 10   & 56.48\%  &  60.83\% \\
            CIFAR100 & 20   & 50.83\%  &  53.54\%  \\
            SVHN     & 5    & 88.91\%  &  90.83\%  \\
            MS-Celeb & 10   & 65.00\%  &  69.20\%  \\
            \bottomrule
            \end{tabular}
            \end{center}
            \label{tbl:old_vs_new}
\end{minipage}  
\end{table}

\textbf{RPS-Net vs. \ours{}:} In terms of computational efficiency, our proposed adaptive path selection algorithm uses less number of FLOPS to gain the same performance as RPS-Net~\cite{rajasegaran2019random}. For example, without the adaptive path selection, RPS-Net achieves 59.93\% on CIFAR-100 with 10 tasks utilizing 5 number of paths. However, with \ours{}, we achieve 59.15\% with $th=2$ and only $3$ paths. This shows a $\sim$40\% improvement in computationally efficiency at inference time. Further, performance-wise \ours{} surpasses \cite{rajasegaran2019random} by a considerable margin on all datasets. As an example, we consider CIFAR-100 with 10 and 20 incremental tasks, the modified RPS-Net with adaptive path selection and path attention surpasses RPS-Net~\cite{rajasegaran2019random} by 4.35\% and 2.71\% respectively. In Table~\ref{tbl:old_vs_new}, we provide more comparisons on the performance gain by \ours{}.

\section{Conclusion}
In real-life setting, learning tasks appear in a sequential order and an autonomous agent must continually increment its existing knowledge. Deep neural networks excel in the cumulative learning setting where all tasks are available all together, but their performance deteriorates for the incremental learning case. In this paper, we propose a scalable approach to class-incremental learning that aims to keep the right balance between previously acquired knowledge and the newly presented tasks. We achieve this using an optimal path selection approach that supports parallelism and knowledge exchange between old and new tasks. Our approach can automatically estimate path capacity with in the network and subsequently decide if a new path is required to continue learning new tasks. Further, a controlling mechanism is introduced to maintain an equilibrium between the stability and plasticity of the learned model. Our approach delivers strong performance gains on MNIST, SVHN, CIFAR-100, MS-Celeb and ImageNet datasets for incremental learning problem.


\ifCLASSOPTIONcaptionsoff
  \newpage
\fi

{\small
\bibliographystyle{IEEEtran}
\bibliography{ref}

\begin{thebibliography}{10}
\providecommand{\url}[1]{#1}
\csname url@samestyle\endcsname
\providecommand{\newblock}{\relax}
\providecommand{\bibinfo}[2]{#2}
\providecommand{\BIBentrySTDinterwordspacing}{\spaceskip=0pt\relax}
\providecommand{\BIBentryALTinterwordstretchfactor}{4}
\providecommand{\BIBentryALTinterwordspacing}{\spaceskip=\fontdimen2\font plus
\BIBentryALTinterwordstretchfactor\fontdimen3\font minus
  \fontdimen4\font\relax}
\providecommand{\BIBforeignlanguage}[2]{{%
\expandafter\ifx\csname l@#1\endcsname\relax
\typeout{** WARNING: IEEEtran.bst: No hyphenation pattern has been}%
\typeout{** loaded for the language `#1'. Using the pattern for}%
\typeout{** the default language instead.}%
\else
\language=\csname l@#1\endcsname
\fi
#2}}
\providecommand{\BIBdecl}{\relax}
\BIBdecl

\bibitem{mccloskey1989catastrophic}
M.~McCloskey and N.~J. Cohen, ``Catastrophic interference in connectionist
  networks: The sequential learning problem,'' in \emph{Psychology of learning
  and motivation}.\hskip 1em plus 0.5em minus 0.4em\relax Elsevier, 1989,
  vol.~24, pp. 109--165.

\bibitem{khan2018guide}
S.~Khan, H.~Rahmani, S.~A.~A. Shah, and M.~Bennamoun, ``A guide to
  convolutional neural networks for computer vision,'' \emph{Synthesis Lectures
  on Computer Vision}, vol.~8, no.~1, pp. 1--207, 2018.

\bibitem{li2018learning}
Z.~Li and D.~Hoiem, ``Learning without forgetting,'' \emph{IEEE transactions on
  pattern analysis and machine intelligence}, vol.~40, no.~12, pp. 2935--2947,
  2018.

\bibitem{rebuffi2017icarl}
S.-A. Rebuffi, A.~Kolesnikov, G.~Sperl, and C.~H. Lampert, ``icarl: Incremental
  classifier and representation learning,'' in \emph{Proceedings of the IEEE
  Conference on Computer Vision and Pattern Recognition}, 2017, pp. 2001--2010.

\bibitem{rusu2016progressive}
A.~A. Rusu, N.~C. Rabinowitz, G.~Desjardins, H.~Soyer, J.~Kirkpatrick,
  K.~Kavukcuoglu, R.~Pascanu, and R.~Hadsell, ``Progressive neural networks,''
  \emph{arXiv preprint arXiv:1606.04671}, 2016.

\bibitem{kirkpatrick2017overcoming}
J.~Kirkpatrick, R.~Pascanu, N.~Rabinowitz, J.~Veness, G.~Desjardins, A.~A.
  Rusu, K.~Milan, J.~Quan, T.~Ramalho, A.~Grabska-Barwinska \emph{et~al.},
  ``Overcoming catastrophic forgetting in neural networks,'' in
  \emph{Proceedings of the national academy of sciences}, vol. 114,
  no.~13.\hskip 1em plus 0.5em minus 0.4em\relax National Acad Sciences, 2017,
  pp. 3521--3526.

\bibitem{kemker2018measuring}
R.~Kemker, M.~McClure, A.~Abitino, T.~L. Hayes, and C.~Kanan, ``Measuring
  catastrophic forgetting in neural networks,'' in \emph{Thirty-second AAAI
  conference on artificial intelligence}, 2018.

\bibitem{he2016deep}
K.~He, X.~Zhang, S.~Ren, and J.~Sun, ``Deep residual learning for image
  recognition,'' in \emph{Proceedings of the IEEE conference on computer vision
  and pattern recognition}, 2016, pp. 770--778.

\bibitem{rajasegaran2019random}
J.~Rajasegaran, M.~Hayat, S.~Khan, F.~S. Khan, and L.~Shao, ``Random path
  selection for incremental learning,'' \emph{Advances in Neural Information
  Processing Systems}, 2019.

\bibitem{abraham2005memory}
W.~C. Abraham and A.~Robins, ``Memory retention--the synaptic stability versus
  plasticity dilemma,'' \emph{Trends in neurosciences}, vol.~28, no.~2, pp.
  73--78, 2005.

\bibitem{hinton2015distilling}
G.~Hinton, O.~Vinyals, and J.~Dean, ``Distilling the knowledge in a neural
  network,'' \emph{arXiv preprint arXiv:1503.02531}, 2015.

\bibitem{Castro_2018_ECCV}
F.~M. Castro, M.~J. Marin-Jimenez, N.~Guil, C.~Schmid, and K.~Alahari,
  ``End-to-end incremental learning,'' in \emph{The European Conference on
  Computer Vision (ECCV)}, September 2018.

\bibitem{zenke2017continual}
F.~Zenke, B.~Poole, and S.~Ganguli, ``Continual learning through synaptic
  intelligence,'' in \emph{Proceedings of the 34th International Conference on
  Machine Learning-Volume 70}.\hskip 1em plus 0.5em minus 0.4em\relax JMLR.
  org, 2017, pp. 3987--3995.

\bibitem{hou2018lifelong}
S.~Hou, X.~Pan, C.~Change~Loy, Z.~Wang, and D.~Lin, ``Lifelong learning via
  progressive distillation and retrospection,'' in \emph{Proceedings of the
  European Conference on Computer Vision (ECCV)}, 2018, pp. 437--452.

\bibitem{kemker2017fearnet}
R.~Kemker and C.~Kanan, ``Fearnet: Brain-inspired model for incremental
  learning,'' \emph{International Conference on Learning Representations},
  2018.

\bibitem{gepperth2016bio}
A.~Gepperth and C.~Karaoguz, ``A bio-inspired incremental learning architecture
  for applied perceptual problems,'' \emph{Cognitive Computation}, vol.~8,
  no.~5, pp. 924--934, 2016.

\bibitem{kamra2017deep}
N.~Kamra, U.~Gupta, and Y.~Liu, ``Deep generative dual memory network for
  continual learning,'' \emph{arXiv preprint arXiv:1710.10368}, 2017.

\bibitem{xiang2019incremental}
Y.~Xiang, Y.~Fu, P.~Ji, and H.~Huang, ``Incremental learning using conditional
  adversarial networks,'' in \emph{Proceedings of the IEEE International
  Conference on Computer Vision}, 2019, pp. 6619--6628.

\bibitem{belouadah2019il2m}
E.~Belouadah and A.~Popescu, ``Il2m: Class incremental learning with dual
  memory,'' in \emph{Proceedings of the IEEE International Conference on
  Computer Vision}, 2019, pp. 583--592.

\bibitem{xiao2014error}
T.~Xiao, J.~Zhang, K.~Yang, Y.~Peng, and Z.~Zhang, ``Error-driven incremental
  learning in deep convolutional neural network for large-scale image
  classification,'' in \emph{Proceedings of the 22nd ACM international
  conference on Multimedia}.\hskip 1em plus 0.5em minus 0.4em\relax ACM, 2014,
  pp. 177--186.

\bibitem{fernando2017pathnet}
C.~Fernando, D.~Banarse, C.~Blundell, Y.~Zwols, D.~Ha, A.~A. Rusu, A.~Pritzel,
  and D.~Wierstra, ``Pathnet: Evolution channels gradient descent in super
  neural networks,'' \emph{arXiv preprint arXiv:1701.08734}, 2017.

\bibitem{lopez2017gradient}
D.~Lopez-Paz \emph{et~al.}, ``Gradient episodic memory for continual
  learning,'' in \emph{Advances in Neural Information Processing Systems},
  2017, pp. 6467--6476.

\bibitem{xie2019exploring}
S.~Xie, A.~Kirillov, R.~Girshick, and K.~He, ``Exploring randomly wired neural
  networks for image recognition,'' \emph{arXiv preprint arXiv:1904.01569},
  2019.

\bibitem{zoph2018learning}
B.~Zoph, V.~Vasudevan, J.~Shlens, and Q.~V. Le, ``Learning transferable
  architectures for scalable image recognition,'' in \emph{Proceedings of the
  IEEE conference on computer vision and pattern recognition}, 2018, pp.
  8697--8710.

\bibitem{pham2018efficient}
H.~Pham, M.~Y. Guan, B.~Zoph, Q.~V. Le, and J.~Dean, ``Efficient neural
  architecture search via parameter sharing,'' \emph{arXiv preprint
  arXiv:1802.03268}, 2018.

\bibitem{Hsu18_EvalCL}
\BIBentryALTinterwordspacing
Y.-C. Hsu, Y.-C. Liu, A.~Ramasamy, and Z.~Kira, ``Re-evaluating continual
  learning scenarios: A categorization and case for strong baselines,'' 2018.
  [Online]. Available: \url{https://arxiv.org/abs/1810.12488}
\BIBentrySTDinterwordspacing

\bibitem{GeM_pooling}
F.~{Radenović}, G.~{Tolias}, and O.~{Chum}, ``Fine-tuning cnn image retrieval
  with no human annotation,'' \emph{IEEE Transactions on Pattern Analysis and
  Machine Intelligence}, vol.~41, no.~7, pp. 1655--1668, July 2019.

\bibitem{kingma2014adam}
D.~P. Kingma and J.~Ba, ``Adam: A method for stochastic optimization,''
  \emph{arXiv preprint arXiv:1412.6980}, 2014.

\bibitem{chaudhry2018riemannian}
A.~Chaudhry, P.~K. Dokania, T.~Ajanthan, and P.~H. Torr, ``Riemannian walk for
  incremental learning: Understanding forgetting and intransigence,'' in
  \emph{Proceedings of the European Conference on Computer Vision (ECCV)},
  2018, pp. 532--547.

\bibitem{aljundi2018memory}
R.~Aljundi, F.~Babiloni, M.~Elhoseiny, M.~Rohrbach, and T.~Tuytelaars, ``Memory
  aware synapses: Learning what (not) to forget,'' in \emph{Proceedings of the
  European Conference on Computer Vision (ECCV)}, 2018, pp. 139--154.

\bibitem{zhang2019class}
J.~Zhang, J.~Zhang, S.~Ghosh, D.~Li, S.~Tasci, L.~Heck, H.~Zhang, and C.-C.~J.
  Kuo, ``Class-incremental learning via deep model consolidation,'' \emph{arXiv
  preprint arXiv:1903.07864}, 2019.

\bibitem{schwarz2018progress}
J.~Schwarz, J.~Luketina, W.~M. Czarnecki, A.~Grabska-Barwinska, Y.~W. Teh,
  R.~Pascanu, and R.~Hadsell, ``Progress \& compress: A scalable framework for
  continual learning,'' \emph{arXiv preprint arXiv:1805.06370}, 2018.

\bibitem{shin2017continual}
H.~Shin, J.~K. Lee, J.~Kim, and J.~Kim, ``Continual learning with deep
  generative replay,'' in \emph{Advances in Neural Information Processing
  Systems}, 2017, pp. 2990--2999.

\bibitem{van2018generative}
G.~M. van~de Ven and A.~S. Tolias, ``Generative replay with feedback
  connections as a general strategy for continual learning,'' \emph{arXiv
  preprint arXiv:1809.10635}, 2018.

\bibitem{DBLP:journals/corr/abs-1802-07569}
\BIBentryALTinterwordspacing
G.~I. Parisi, R.~Kemker, J.~L. Part, C.~Kanan, and S.~Wermter, ``Continual
  lifelong learning with neural networks: {A} review,'' \emph{CoRR}, vol.
  abs/1802.07569, 2018. [Online]. Available:
  \url{http://arxiv.org/abs/1802.07569}
\BIBentrySTDinterwordspacing

\bibitem{miller1995genetic}
B.~L. Miller, D.~E. Goldberg \emph{et~al.}, ``Genetic algorithms, tournament
  selection, and the effects of noise,'' \emph{Complex systems}, vol.~9, no.~3,
  pp. 193--212, 1995.

\end{thebibliography}
}

%



\end{document}